\documentclass{article}

\usepackage{PRIMEarxiv}

\usepackage[utf8]{inputenc} 
\usepackage[T1]{fontenc}    
\usepackage{hyperref}       
\usepackage{url}            
\usepackage{booktabs}       
\usepackage{amsfonts}       
\usepackage{nicefrac}       
\usepackage{microtype}      
\usepackage{fancyhdr}       
\usepackage{graphicx}       
\graphicspath{{media/}}     
\usepackage{graphicx}
\usepackage{multirow} 
\usepackage{multicol} 
\usepackage{colortbl}
\usepackage{array}
\usepackage{makecell}
\usepackage{amsmath}
\usepackage{amssymb}
\usepackage[export]{adjustbox}
\usepackage[table]{xcolor}
\usepackage{subcaption}
\usepackage[numbers]{natbib}
\usepackage{hyperref}

\definecolor{lightgreen}{RGB}{230,255,230}
\definecolor{lightred}{RGB}{255,230,230}

\usepackage{booktabs}
\usepackage{tabularx}
\newcolumntype{Y}{>{\centering\arraybackslash}X}
\usepackage{multirow}

\usepackage{stackengine}

\title{Your Image Generator Is Your New Private Dataset
}

\author{
  Nicolò Francesco Resmini, Eugenio Lomurno, Cristian Sbrolli, and Matteo Matteucci \\
  Politecnico di Milano \\
  Department of Electronics, Information and Bioengineering\\
  Via Ponzio 34/5, 20133 Milan, Italy\\
  \texttt{nicolofrancesco.resmini@mail.polimi.it} \\
  \texttt{eugenio.lomurno@polimi.it} \\
  \texttt{cristian.sbrolli@polimi.it} \\
  \texttt{matteo.matteucci@polimi.it} \\
}

\begin{document}
\maketitle

\begin{abstract}
Generative diffusion models have emerged as powerful tools to synthetically produce training data, offering potential solutions to data scarcity and reducing labelling costs for downstream supervised deep learning applications. However, effectively leveraging text-conditioned image generation for building classifier training sets requires addressing key issues: constructing informative textual prompts, adapting generative models to specific domains, and ensuring robust performance. This paper proposes the Text-Conditioned Knowledge Recycling (TCKR) pipeline to tackle these challenges. TCKR combines dynamic image captioning, parameter-efficient diffusion model fine-tuning, and Generative Knowledge Distillation techniques to create synthetic datasets tailored for image classification. The pipeline is rigorously evaluated on ten diverse image classification benchmarks. The results demonstrate that models trained solely on TCKR-generated data achieve classification accuracies on par with (and in several cases exceeding) models trained on real images. Furthermore, the evaluation reveals that these synthetic-data-trained models exhibit substantially enhanced privacy characteristics: their vulnerability to Membership Inference Attacks is significantly reduced, with the membership inference AUC lowered by 5.49 points on average compared to using real training data, demonstrating a substantial improvement in the performance-privacy trade-off. These findings indicate that high-fidelity synthetic data can effectively replace real data for training classifiers, yielding strong performance whilst simultaneously providing improved privacy protection as a valuable emergent property. The code and trained models are available in the accompanying \href{https://github.com/NicoloResmini/TCKR}{open-source repository}.
\end{abstract}


\section{Introduction}
\label{sec:introduction}
\noindent Generative models have substantially transformed machine learning, particularly in computer vision where text-to-image diffusion models demonstrate remarkable capabilities in synthetic data creation. These developments extend beyond artistic applications, offering solutions to persistent challenges in machine learning, including data scarcity, privacy concerns, and training dataset imbalances~\cite{bansal2022systematic}. Traditional image classification approaches have relied on large-scale annotated datasets that present significant limitations: intensive human labelling effort, potential privacy violations, and inherent collection biases. In specialised domains such as medical imaging or industrial inspection, obtaining sufficient annotated examples is often prohibitively expensive or practically impossible~\cite{litjens2017survey}. While synthetic data generation offers an alternative without direct privacy implications, earlier approaches typically produced images with poor visual fidelity and limited diversity, resulting in inadequate real-world performance~\cite{figueira2022survey}.

\noindent Text-conditioned diffusion models like Stable Diffusion~\cite{rombach2022high} have substantially narrowed this utility gap. Pre-trained on diverse image-text pairs, these models can generate images with unprecedented detail and semantic richness guided by textual descriptions. Unlike previous generative approaches that require class labels or exemplar images, text-conditioned models create varied samples based solely on linguistic descriptions, potentially capturing subtle visual characteristics that define different classes. Despite these advances, key challenges persist in using text-conditioned diffusion models to build classification-optimised synthetic datasets. Effective textual prompt selection is crucial, as basic class names often fail to capture the natural intra-class variations found in real datasets. Additionally, ensuring that generated images contain the correct semantic properties for classification requires careful domain adaptation of the generative model. Finally, questions remain about how synthetic data influences model privacy~-- particularly regarding vulnerability to Membership Inference Attacks~\cite{shokri2017membership}.

\noindent This paper introduces \emph{Text-Conditioned Knowledge Recycling (TCKR)}, a comprehensive pipeline that addresses these challenges by combining advanced text-conditioned image generation with efficient generator adaptation and classifier knowledge transfer. By integrating dynamic captioning with BLIP-2~\cite{li2023blip} to produce instance-specific prompts, parameter-efficient fine-tuning (LoRA) of the diffusion model, and Generative Knowledge Distillation~\cite{lomurno2025synthetic} to craft more informative labels, TCKR produces synthetic datasets that maintain high utility for classification while simultaneously strengthening privacy protection. This research investigates whether synthetically generated datasets can achieve classification performance comparable to real data while improving resistance to privacy attacks. In this work it is explored how different text-conditioning strategies impact performance, examine the relationship between synthetic dataset size and classification accuracy, and analyse the trade-offs between model accuracy and privacy.

\noindent The contributions of this paper include:
\begin{itemize}
    \item The TCKR pipeline combining text-conditioned diffusion models, parameter-efficient model adaptation, and Generative Knowledge Distillation to generate highly informative synthetic training datasets for image classification.
    \item A dynamic captioning strategy using BLIP-2 that captures instance-specific visual attributes of images, improving the semantic quality and diversity of generated synthetic data.
    \item Empirical evaluation on 10 diverse datasets demonstrating that classifiers trained exclusively on TCKR-synthesised data can achieve accuracy comparable to -- and in several cases exceeding -- models trained on real data, while exhibiting substantially enhanced privacy, i.e. a lower membership inference vulnerability and a better accuracy-privacy trade-off.
\end{itemize}

\section{Related Works}
\label{sec:related_works}
\noindent In order to produce synthetic datasets that can be used to train models capable of performing effectively on real data, it is crucial to resort to generative architectures that guarantee high-quality image synthesis. The principal families of models that have stood out in this task are the Generative Adversarial Networks (GAN) and the Denoising Diffusion Probabilistic Models (DDPM)~\cite{goodfellow2014generative, ho2020denoising}. Although they operate on the basis of different mechanisms, both families can be conditioned in various ways, making it possible to produce images that align with specific domains or visual concepts.

\subsection{Conditioning Methods}\label{subsec:conditioning_methods}
\noindent Shortly after the introduction of the GAN, a conditional formulation was proposed, allowing the generation process to be guided by additional information provided to both generator and discriminator~\cite{mirza2014conditional}. In particular, conditioning on class labels has demonstrated benefits through scaling the training set and through the truncation of the variance in the noise input, an approach that can help enhance image quality~\cite{brock2018large}.

\noindent More recently, the possibility of conditioning generation on text prompts has been widely explored. One line of research focuses on improving the quality of the textual representation used for generation. For instance, Ku \textit{et al.} proposed training a regressor that produces more precise text-conditioned vectors, facilitating fine control over minor features in the generated images~\cite{ku2023textcontrolgan}. Another work by Tao \textit{et al.} exploited CLIP's broad understanding of visual scenes, combining a CLIP-based discriminator with a CLIP-enhanced generator to reduce training time while improving the synthesized output~\cite{tao2023galip, radford2021learning}.

\noindent Diffusion-based solutions also employ textual prompts and often adopt guidance strategies to balance fidelity and diversity. Nichol \textit{et al.} demonstrated that classifier-free guidance, achieved by blending the model's predictions with and without text conditioning, can outperform approaches relying on CLIP guidance~\cite{nichol2021glide}. Stable Diffusion itself incorporates a parameter named Guidance Scale to govern the adherence of generated images to the text prompt~\cite{rombach2022high}. Furthermore, several works explored rewriting prompts with large language models to enhance semantic alignment~\cite{yang2024mastering, saharia2022photorealistic, betker2023improving}, while other approaches leveraged newly introduced tokens in the text embedding space to teach the model novel concepts or styles, leading to a higher variety of generated outcomes~\cite{ruiz2023dreambooth, gal2022image}.

\subsection{Learning from Synthetic Data}
\label{subsec:learning_from_synthetic_data}
\noindent Recently, it has been shown that a retroactive process may be at work, by which the quality of individual images, now increasingly present on the web, and their often indistinguishability to the human eye from those created by humans themselves, is increasing. It has been shown that generative models trained on images produced by other generative models tend to degrade from generation to generation until they collapse~\cite{shumailov2024ai}. As it is very complex to detect the nature of such data, this is undoubtedly a major open problem.

\noindent Then, although both GAN and DDPM frameworks allow for the generation of single images with high perceptual realism, a key challenge lies in creating entire synthetic datasets that exhibit enough diversity to train downstream models effectively -- such as classification or generative models. 

\noindent Sariyildiz \textit{et al.} demonstrated that by adopting a minimal form of prompt engineering, combined with a reduced guidance scale for Stable Diffusion, it is possible to match or closely approach the performance attained with real data~\cite{sariyildiz2023fake}. Indeed, further studies highlighted how the value of the guidance scale is crucial for synthetic images to be genuinely beneficial for downstream classification tasks~\cite{lomurno2024stable}. Furthermore, in some cases, simply increasing the size of the synthetic dataset can exceed the performance obtained from real data alone, underscoring the importance of sample diversity in synthetic learning. At the same time, it has been shown how the expansion of the variance in the noise input leads GANs to produce more informative synthetic datasets even when composed of images of lesser visual quality~\cite{lampis2023bridging}.

\noindent Another research direction aimed at varying the composition of the textual prompt in a manner that is independent of both the target dataset and the classification model, with the goal of further boosting diversity. Shipard \textit{et al.} proposed a “Bag of Tricks” consisting of multiple predefined prompt modifications that can be combined to generate a larger and more varied synthetic dataset~\cite{shipard2023diversity}. On the other hand, Lei \textit{et al.} investigated the use of image captioners to create prompts composed of captions concatenated with class labels, thus clarifying the distinction between foreground and background elements~\cite{lei2023image}. In the same vein, other pipelines employed captioners to propose a set of possible prompts, and then resorted to CLIP similarity to select the most fitting description for each image~\cite{li2024semantic}.

\noindent In parallel, techniques were proposed to enrich the synthetic data after their generation to make them more useful in downstream classification tasks. The Generative Knowledge Distillation technique makes it possible to exploit a model trained on real data -- called Teacher Classifier -- to create soft labels for the generated images to be used as a training set for a model to be trained only from these -- called Student Classifier~\cite{lomurno2025synthetic}. The use of this training technique -- which is part of the Knowledge Recycling pipeline, involving the tuning of parameters such as the regeneration of the synthetic dataset, the expansion of the generation standard deviation and the increase of the cardinality of the synthetic dataset with respect to the real one -- proved to be extremely effective in increasing the performance of the Student Classifier as well as its resilience to inference attacks, in local as well as in federated contexts~\cite{lomurno2025federated}.

\subsection{Privacy Threats and Synthetic Data Defences}
\label{subsec:privacy_threats_and_synthetic_data_defences}
\noindent The landscape of deep learning security encompasses various privacy threats, including model inversion, extraction attacks, and data poisoning~\cite{fredrikson2015model, tramer2016stealing, biggio2012poisoning}. Among these, Membership Inference Attacks~\cite{shokri2017membership} emerge as a particularly critical preliminary threat, as their success often facilitates subsequent privacy breaches. These sophisticated attacks enable adversaries to determine whether specific samples were used in model training, potentially compromising sensitive information even in black-box settings where access to model parameters is restricted. Advanced techniques such as the Likelihood Ratio Attack (LiRA)~\cite{carlini2022membership} leverage shadow models to achieve remarkable inference accuracy, demonstrating the evolving sophistication of privacy threats. While traditional defences such as Differential Privacy~\cite{abadi2016deep} offer theoretical guarantees through noise injection during training, they often introduce substantial compromises to model utility~\cite{lomurno2022utility}. In this context, synthetic data emerge as a promising defence strategy. By training models on generated rather than real samples, organizations can effectively reduce MIA risks while maintaining performance levels~\cite{hu2022defending, lomurno2025synthetic, lomurno2025federated}. This approach represents a modern solution to the persistent challenge of balancing privacy requirements with practical utility in deep learning applications, offering a pathway to robust model development without direct exposure of sensitive training data.


\section{Method}\label{sec:method}
\begin{figure}
    \centering
    \includegraphics[width=1\textwidth]{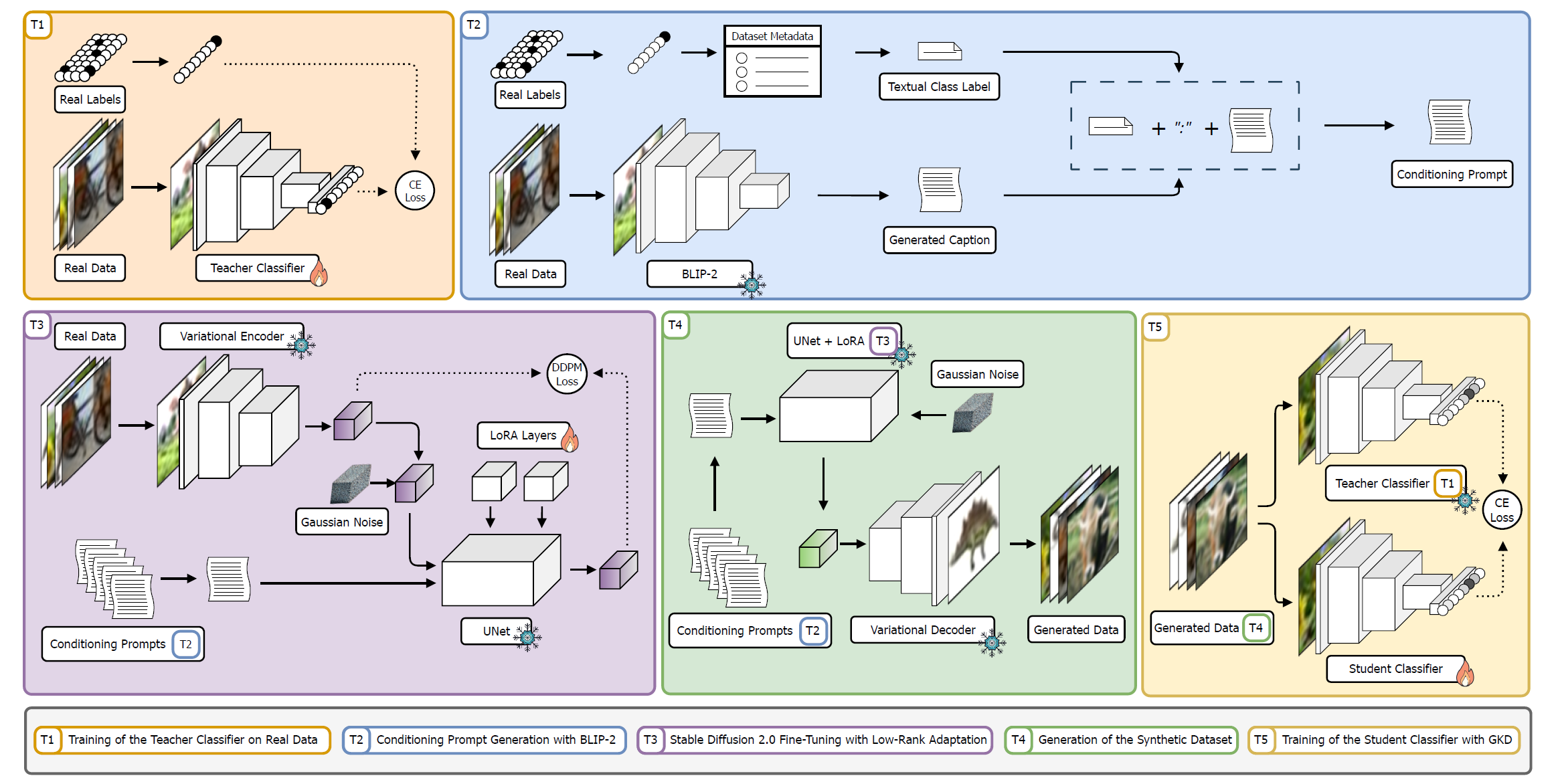}
    \caption{Graphical abstract of the Text-Conditioned Knowledge Recycling (TCKR) pipeline summarising the entire process employed to generate synthetic datasets. Initially, for each image, a specific caption is produced using the BLIP‑2 model. Thereafter, the generator, based on Stable Diffusion 2.0, is adapted to the target domain using the LoRA technique and is conditioned via prompts that combine the class name with the caption (formatted as “n: c”). Finally, the Generative Knowledge Distillation transfers the knowledge from the teacher classifier to the student classifier, thereby enabling the development of models that achieve high classification performance whilst ensuring enhanced privacy protection.}
    \label{fig:graphical_abstract}
\end{figure}

\noindent This section presents the contribution of this work, which consists of the Text-Conditioned Knowledge Recycling (TCKR) pipeline for generating synthetic datasets with high resistance to Membership Inference Attacks and high information level for downstream image classification tasks. An overview of the entire TCKR pipeline is depicetd in Figure~\ref{fig:graphical_abstract}.

\subsection{Generator, Adaptation and Prompting}\label{subsec:generator_adaptation_and_prompting}
\noindent The TCKR pipeline begins with the selection of a Generator, used for the creation of the synthetic dataset, and its adaptation to the target dataset. In contrast to the experiments with which the Knowledge Recycling~\cite{lomurno2025synthetic} pipeline was presented~-- relying on a variation of the BigGAN-Deep~\cite{brock2018large} model trained from scratch~-- in this work the Stable Diffusion 2.0~\cite{rombach2022high} model pre-trained on LAION-5B~\cite{schuhmann2022laion} is used as the Generator~-- which was chosen following a comparison with other models of the same family and whose results are shown in Appendix~A. The generation parameters used for Stable Diffusion 2.0 differ from the original implementation in favour of those suggested by Sariyildiz \textit{et al.}~-- Unconditional Guidance Scale = 2 and Inference Steps = 20, with the latter chosen to optimize computation time while remaining close to their recommended range of 25-50 steps~-- as they are more effective in generating more heterogeneous and informative datasets~\cite{sariyildiz2023fake}. This model is adapted to the target dataset using the Low-Rank Adaptation (LoRA) technique~\cite{hu2021lora}.

\noindent LoRA is a lightweight fine-tuning technique that introduces additional rank-decomposed parameters without significantly increasing the total number of trainable parameters. Formally, given \(\theta\) the set of parameters of the basic Stable Diffusion 2.0 model, LoRA adds a low-dimensional update \(\Delta \theta\) specific to the new domain, thus obtaining:
\[
    \theta' = \theta + \Delta \theta, \qquad \Delta \theta = U \, V^\top,
\]
where \(U\) and \(V\) are low rank matrices, both learned during the fitting process. This approach preserves much of the expressive capacity of the original model, while making the fit much more parameter-efficient. In the experiments conducted, 3 epochs of adaptation were used, employing mixed precision and fixing the image size at \(224 \times 224 \). For further details and insights, please refer to Appendix~B.

\noindent As this is text-conditioned generation~-- rather than label-conditioned as in the KR pipeline~-- a key aspect is the choice of the textual prompt with which the Generator is adapted and with which the generation process is subsequently conditioned.
Whereas previous work is often based on the use of the class name alone or in combination with a fixed class description, such as that obtained from the lexical database WordNet~\cite{sariyildiz2023fake, miller1995wordnet}, in the TCKR pipeline captions are generated dynamically for each image in the training set and prior to adaptation via the BLIP-2 model~\cite{li2023blip}.
The pre-trained BLIP-2 implementation employed in this work leverages the OPT large language model~\cite{zhang2022opt} with $2.7$ billion parameters, selected from the Hugging Face~\cite{wolf2020transformers} repository\footnote{\scriptsize\url{https://huggingface.co/Salesforce/blip2-opt-2.7b}} for its computational efficiency. This model was applied without dataset-specific fine-tuning, with caption generation controlled solely through the \texttt{max\_new\_tokens=20} parameter.

\noindent Specifically, let $D$ be an image dataset consisting of labelled samples, where each sample $(x, y)$ consists of an image $x$ and an integer label belonging to class $y$ and associated with a class name $n$. For each randomly selected sample $(x, y)$ in $D$, the BLIP-2 model is used to generate a caption $c$ corresponding to image $x$. The adaptation prompt used consists of combining class name $n$ with this caption $c$ separated by a colon:
\begin{center}
“n: c".
\end{center}
This prompt is the result of a study comparing the effectiveness of different prompting techniques, shown in detail in Appendix~C.

\subsection{Generation and Evaluation of Dataset Synthesis}\label{subsec:generation_and_evaluation_of_dataset_synthesis}
\noindent Once the Generator has been adapted, it is used to generate synthetic datasets. During the generation phase, the same prompts used in the adaptation process are used, taken from the real dataset. In order to evaluate the advantage offered by the possibility of generating an unlimited number of images, in this study are produced datasets with cardinality up to 20 times that of the real dataset, reusing the same prompts but with different generation seeds.

\noindent According to the KR pipeline, the evaluation of the information inherent in such datasets is calculated as the Accuracy achieved by models trained exclusively on such synthetic data and evaluated on a real test set~-- i.e. by means of the metric known as the Classification Accuracy Score~\cite{ravuri2019classification}. This Accuracy is then compared with the one achievable by traditional training~-- i.e. by learning the real dataset and performing the evaluation on the real test set.

\noindent To ensure a fair evaluation, classifiers must share the same architecture, be trained with identical methods and have sufficient capabilities to detect variations in the dataset content. POMONAG~\cite{lomurno2024pomonag}, a neural architecture search algorithm capable of generating pre-trained architectures on ImageNet1K~\cite{deng2009imagenet} belonging to the MobileNetV3~\cite{howard2019searching} search space, is used for this purpose. Once this architecture is identified, it is used to build the Teacher Classifier, by fine-tuning it on the real dataset to adapt it specifically to the classification task under consideration~-- details on the hyper-parameters used are presented in Appendix~C.

\noindent At this point, the Generative Knowledge Distillation (GKD)~\cite{lomurno2025synthetic} technique~-- a fundamental step within the KR pipeline~-- is applied to replace the hard labels representing the image classes in the generated datasets. From a practical point of view, the Teacher Classifier is used to evaluate the generated images and produce soft labels in the form of logits. These probabilistic labels have been shown to be significantly more informative than binary labels, as they are able to capture uncertainties and correlations between classes.
Once completed, these synthetic datasets are ready to be used in the training phase of the Student Classifiers, which are built with the same architecture and adapted with the same fine-tuning strategy as the Teacher Classifier, but each on the synthetic datasets of different cardinality just defined, and finally evaluated on the real test set in order to obtain a robust and fair Accuracy comparison.


\subsection{Membership Inference Attack}\label{subsec:MIA}
\noindent
As a final step in the TCKR pipeline, the Student Classifier is tested for resilience to Membership Inference Attacks (MIA), which are methods used to determine whether or not certain data samples are part of the model's training set. In this setting, the Student Classifier does not have direct access to sensitive training examples~-- they are only used to train the Generator and the Teacher Classifier. Therefore, the goal of this phase is to determine whether the MIA can reveal the data used to train the Generator by probing the Student Classifier, and to compare the results with an equivalent attack on the Teacher Classifier.

\begin{table*}[t]
    \centering
    \setlength{\tabcolsep}{3pt}
    \caption{Comprehensive characteristics of datasets used in the experimental evaluation. The table details each dataset's semantic domain, training and test set sizes, number of classes, and training samples per class ($mean \pm std$).}
    \footnotesize
    \begin{tabularx}{\textwidth}{lc*{3}{Y}cr}
        \toprule
        \textbf{Dataset} & \textbf{Topic} & \textbf{Train Size} & \textbf{Test Size} & \textbf{Classes} & \textbf{Training Samples per Class} \\
        \midrule
        CIFAR10~\cite{krizhevsky2009learning}         & Animals \& Objects & 50,000 & 10,000 & 10 & 5000 $\pm$ 0\\
        CIFAR100~\cite{krizhevsky2009learning}        & General            & 50,000 & 10,000 & 100 & 500 $\pm$ 0\\
        Oxford-IIIT-Pet~\cite{parkhi2012cats}           & Cats \& Dogs       & 6,281  & 1,109  & 37 & 170 $\pm$ 1\\
        TinyImageNet~\cite{le2015tiny}                 & General            & 100,000 & 10,000 & 200 & 500 $\pm$ 0\\
        StanfordCars~\cite{krause20133d}               & Cars               & 8,144  & 8,041  & 196 & 42 $\pm$ 4\\
        Food101~\cite{bossard2014food}                  & Food               & 75,750 & 25,250 & 101 & 750 $\pm$ 0\\
        STL10~\cite{coates2011analysis}                & Animals \& Objects & 5,000  & 8,000  & 10 & 500 $\pm$ 0\\
        Imagewoof~\cite{howard2020fastai}               & Dogs               & 9,025  & 3,929  & 10 & 902 $\pm$ 114\\
        Imagenette~\cite{howard2020fastai}              & Objects            & 9,469  & 3,925  & 10 & 947 $\pm$ 35\\
        Caltech101~\cite{fei2004learning}              & General            & 3,060  & 6,084  & 102 & 30 $\pm$ 0\\
        \bottomrule
    \end{tabularx}
    \label{tab:datasets_characteristics}
\end{table*}

\noindent The attack is carried out by employing the Likelihood Ratio Attack (LiRA) proposed by Carlini \textit{et al.}~\cite{carlini2022membership}. This state-of-the-art framework recasts membership inference as a hypothesis testing challenge, distinguishing between two distributions: one in which the target instance was part of training, and one in which it was not. The implementation proceeds as follows:

\begin{enumerate}
    \item A total of 256 shadow models are trained, adopting the minimal architecture belonging to the MobileNetV3 search space~-- for efficiency~-- and the same training procedure used for both the Student Classifier and the Teacher Classifier~-- except for the batch size increased to 1024 to speed up the training procedure.
    \item The test dataset is subdivided using a 50/10/40 split to form training, validation, and test sets for the shadow models, ensuring each shadow model receives a unique partition.
    \item For each instance $(x,y)$, the model’s confidence is recorded in logit form:
    \vspace{-2.5pt}
    \[
    \phi(p) = \log\left(\frac{p}{1-p}\right), \quad \text{where} \quad p = f(x)_y.
    \]
    \item Two Gaussian distributions, $\tilde{Q}_{in}$ and $\tilde{Q}_{out}$, are estimated to represent the logit-based confidences when the example is, respectively, included in or excluded from training.
    \item Both an online and an offline variant of the attack are employed:
    \begin{itemize}
        \item \textit{Online}: the means ($\mu_{in}$, $\mu_{out}$) and variances ($\sigma^2_{in}$, $\sigma^2_{out}$) of the in- and out-distributions are directly computed.
        \item \textit{Offline}: only $\mu_{out}$ and $\sigma^2_{out}$ are estimated, enabling a one-sided hypothesis test.
    \end{itemize}
    \item A comparison is conducted between a global and a per-example variance estimation approach, selecting the most effective option for each scenario.
\end{enumerate}

\noindent Ultimately, the likelihood ratio of these two distributions is used to judge membership:
\vspace{-2.5pt}
\[\Lambda = \frac{p(\phi(f(x)_y) | \mathcal{N}(\mu_{in}, \sigma^2_{in}))}{p(\phi(f(x)_y) | \mathcal{N}(\mu_{out}, \sigma^2_{out}))},\]
\vspace{-2.5pt}

\noindent where higher values of $\Lambda$ indicate a greater probability of membership. The same procedure is performed on both the Teacher Classifier and the Student Classifier, leveraging LiRA’s capability to achieve significant true-positive rates at very low false-positive rates, which is a critical factor in privacy audits.

\noindent Two main metrics are used to gauge resilience to MIAs. The first metric is the Area Under the ROC Curve (AUC), often employed to quantify this type of attack. The second metric is the Accuracy Over Privacy (AOP)~\cite{lomurno2023discriminative}, capturing the trade-off between predictive performance -- measured by test accuracy -- and robustness to Membership Inference Attacks.


\section{Experimental Setup}
\label{sec:experimental_setup}

\noindent This section provides a comprehensive description of the experimental environment employed to implement and evaluate the Text-Conditioned Knowledge Recycling (TCKR) pipeline. All experiments are conducted using an NVIDIA Quadro RTX 6000 GPU, which provides the necessary computational resources for efficiently processing and analysing the datasets.

\noindent Experiments were conducted across a diverse collection of image datasets, each with distinct characteristics that allowed for evaluation of the approach across various domains and data distributions. Table~\ref{tab:datasets_characteristics} presents a detailed overview of these datasets, highlighting their variations in size, domain specificity, and sample availability per class.

\noindent It is important to note that for several datasets (specifically Oxford-IIIT-Pet, Imagewoof, Imagenette, and Caltech101), multiple versions exist in the literature with different train/test partitioning schemes. In this work, the versions with the characteristics detailed in Table~\ref{tab:datasets_characteristics} were specifically utilized. Furthermore, for both Imagewoof and Imagenette datasets, the \textit{full size} resolution variant was selected among the three available options (\textit{full size}, \textit{320px}, and \textit{160px}), to maximize the available visual information. In the case of Caltech101, the 102-class variant was employed, which incorporates the additional \textit{Background} class alongside the 101 standard object classes.

\noindent To ensure uniform processing across all datasets and facilitate compatibility with the model architecture, a consistent preprocessing pipeline was implemented. Initially, all images were converted to tensor format and subsequently rescaled to a standardized resolution of $224\times224$ pixels using bicubic interpolation~\cite{keys1981cubic}. This transformation was essential to maintain compatibility with the default input dimensions required by the downstream classifier extracted from MobileNetV3~\cite{howard2019searching}. Bicubic interpolation was specifically selected for the resizing operation because it computes new pixel values based on the 16 nearest pixels in the original image, yielding superior visual quality compared to simpler methods such as nearest neighbour or bilinear interpolation~\cite{han2013comparison, patel2013review}.

\noindent Following the resizing procedure, all images underwent normalization using dataset-specific RGB mean and standard deviation values to standardize the input distribution. The complete sequence of transformations and the augmentation pipeline employed during the training of all Classifiers is thoroughly documented in Appendix~C, while for the generator adaptation via LoRA, a different processing strategy is implemented as detailed in Appendix~B.






\section{Results and Discussion}

\noindent This section provides a comprehensive evaluation of the proposed TCKR pipeline, focusing on its impact on downstream classification performance and on privacy protection (measured as resilience against Membership Inference Attacks).

\subsection{Classification Accuracy Score Evaluation}

\begin{figure*}[t]
    \centering
    \includegraphics[width=\textwidth]{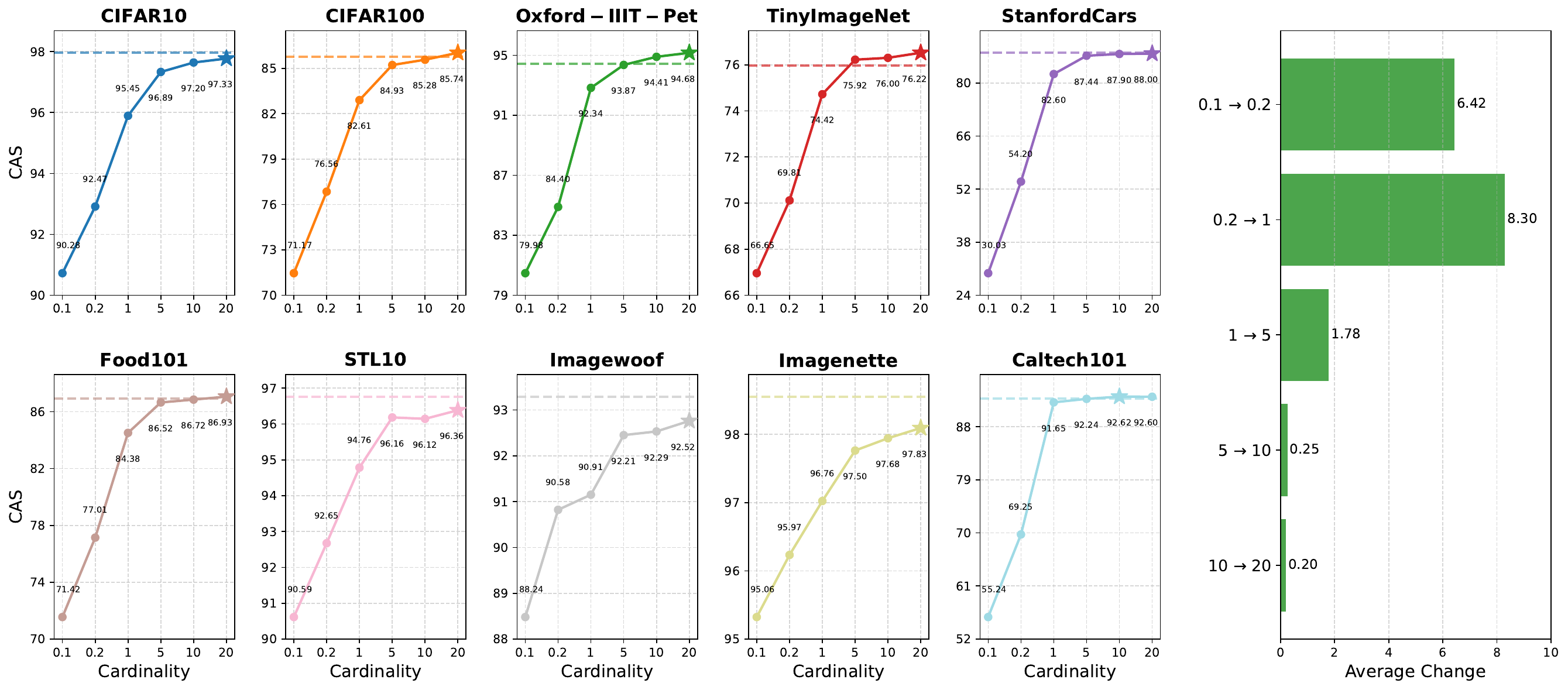}
    \caption{\textbf{Classification Accuracy Score (CAS)} of the Student Classifier for different synthetic dataset cardinalities (left panel). The star marker indicates the Student’s peak CAS on each dataset. The right panel shows the average CAS improvement observed when increasing the synthetic dataset size from one cardinality to the next, averaged across all datasets. For reference, each horizontal dashed line denotes the performance of the corresponding Teacher Classifier.}
    \label{fig:cas_vs_cardinality}
\end{figure*}

\noindent The analysis first evaluates how the size (cardinality) of the synthetic dataset affects classification performance. In these experiments, the synthetic training set size varies from 0.1$\times$ to 20$\times$ the number of images in the corresponding real training set, and the resulting Classification Accuracy Score (CAS) of the Student Classifier is measured. \ref{fig:cas_vs_cardinality} presents the CAS results across this range of cardinalities for all datasets, highlighting overall trends as well as dataset-specific patterns.

\noindent Increasing the synthetic dataset cardinality consistently leads to improved CAS across nearly all datasets. In fact, the highest cardinality (20$\times$) yields the best accuracy in 9 out of 10 benchmarks. The sole exception is Caltech101, where the 10$\times$ synthetic set slightly outperforms 20$\times$ (92.62 vs. 92.60 CAS), though the overall upward trend remains. The magnitude of the accuracy gain varies significantly between datasets. The most pronounced improvement is observed on StanfordCars, where CAS rises from 30.03 at 0.1$\times$ to 88.00 at 20$\times$ – an increase of nearly 58 percentage points. This suggests that complex, fine-grained classification tasks (such as distinguishing among many car models) benefit greatly from larger synthetic training sets. By contrast, datasets that already achieve high accuracy with small synthetic sets (e.g. Imagewoof or STL10) exhibit more modest absolute gains, although their performance still consistently improves with increasing cardinality.

\noindent Notably, the benefit of adding more synthetic data diminishes at very high cardinalities. For instance, on CIFAR100 the CAS jumps by 6.05 points when increasing the synthetic set from 0.2$\times$ to 1$\times$, but by only 0.46 points from 10$\times$ to 20$\times$. This diminishing return – illustrated by the average improvements in \ref{fig:cas_vs_cardinality} ~-- indicates that while more synthetic data generally boosts accuracy, the incremental gain per additional data unit becomes smaller at extreme scales. 

\noindent A correlation exists between the amount of real training data per class (used for diffusion model fine-tuning via LoRA) and the achievable CAS. As shown in \ref{tab:datasets_characteristics}, datasets with a larger number of real samples per class (for example, CIFAR10 with 5,000 images per class) tend to reach higher CAS values at all synthetic data scales. This suggests that the diversity and richness of the real data used during generator adaptation influence the quality of synthetic images and thereby the effectiveness of the synthetic training. However, this correlation is not strict~--~factors such as task complexity and inter-class similarity also impact the classifier's performance on synthetic data.

\noindent Overall, these results demonstrate the potential of leveraging large\nobreakdash-scale synthetic data to enhance classification accuracy across diverse domains. The TCKR pipeline can generate a virtually unlimited pool of training images, and the experiments show that classifier performance continues to improve even up to 20$\times$ the original dataset size. This finding is especially valuable in scenarios where real training data are scarce or costly, as it offers a scalable alternative for improving models without additional real samples. Moreover, in several cases the Student Classifier trained on synthetic data approaches or even exceeds the accuracy of the Teacher Classifier trained on real data. In fact, in five out of the ten datasets, the Student's CAS slightly surpasses the Teacher's accuracy. Importantly, many of these tasks (such as CIFAR100, Oxford-IIIT-Pet, TinyImageNet, STL10, Imagewoof, Imagenette) show no clear performance plateau even at 20$\times$ synthetic data, suggesting that further increasing the synthetic dataset size could yield additional accuracy gains.

\subsection{Privacy Evaluation}
\label{results_privacy}

\begin{figure*}[t]
    \centering
    \includegraphics[width=0.98\textwidth]{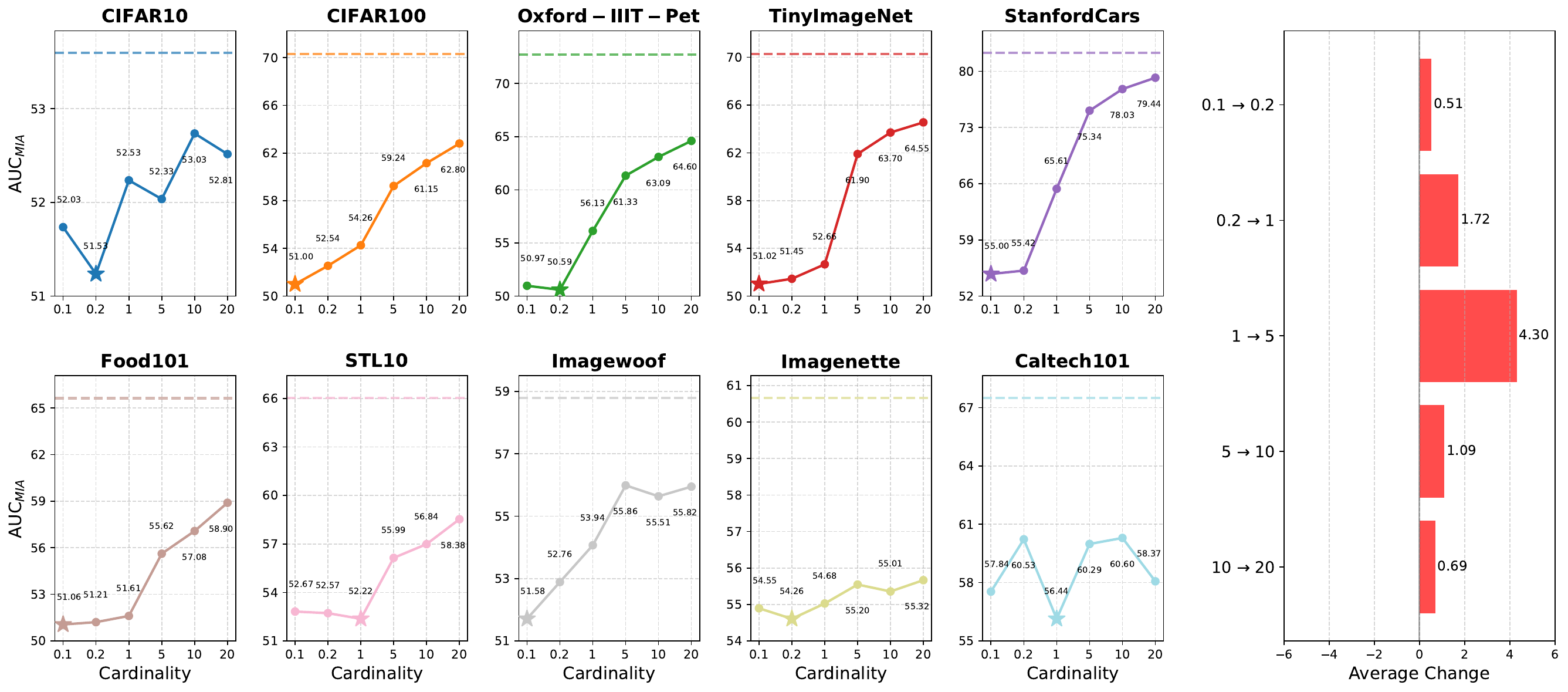}
    \caption{\textbf{Area Under the ROC Curve of the MIA (AUC$_{MIA}$)} for Student Classifiers trained on synthetic datasets of various cardinalities (left panel). The star marker indicates the Student’s lowest AUC$_{MIA}$ (best privacy) achieved. The right panel shows the average AUC$_{MIA}$ increase between successive cardinality levels, averaged across all datasets. For each dataset, the horizontal dashed line represents the AUC$_{MIA}$ of the Teacher Classifier.}
    \label{fig:auc_vs_cardinality}
\end{figure*}

\begin{figure*}[!t]
    \centering
    \includegraphics[width=0.98\textwidth]{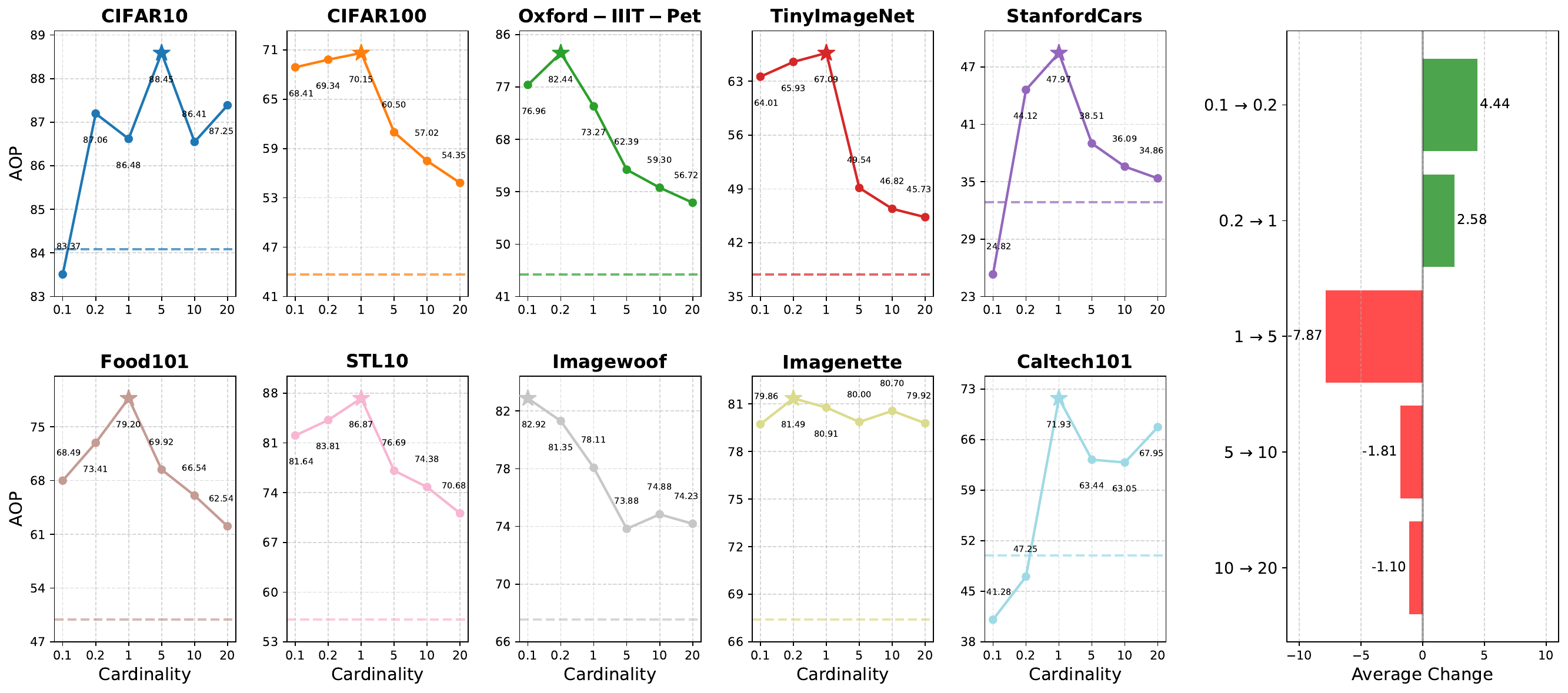}
    \caption{\textbf{Accuracy Over Privacy (AOP)} scores for Student Classifiers at different synthetic dataset cardinalities (left panel), with star markers indicating each Student’s highest AOP. The right panel shows the average AOP change between successive cardinality increases across all datasets. For each dataset, the horizontal dashed line represents the AOP of the corresponding Teacher Classifier.}
    \label{fig:aop_vs_cardinality}
\end{figure*}

\noindent Following the methodology described in Section~3.3, a thorough privacy assessment is conducted next. Here classifiers trained on synthetic datasets of varying cardinalities under Membership Inference Attacks are examined, extending the performance evaluation with a privacy perspective. Specifically, the LiRA Membership Inference Attack~\cite{carlini2022membership} is used against each trained classifier. \ref{fig:auc_vs_cardinality,fig:aop_vs_cardinality} present the results in terms of two complementary privacy metrics: the Area Under the ROC Curve for the Membership Inference Attack (AUC$_{MIA}$) and the Accuracy Over Privacy (AOP) score.



\begin{figure*}[t]
    \centering
    \includegraphics[width=0.95\textwidth]{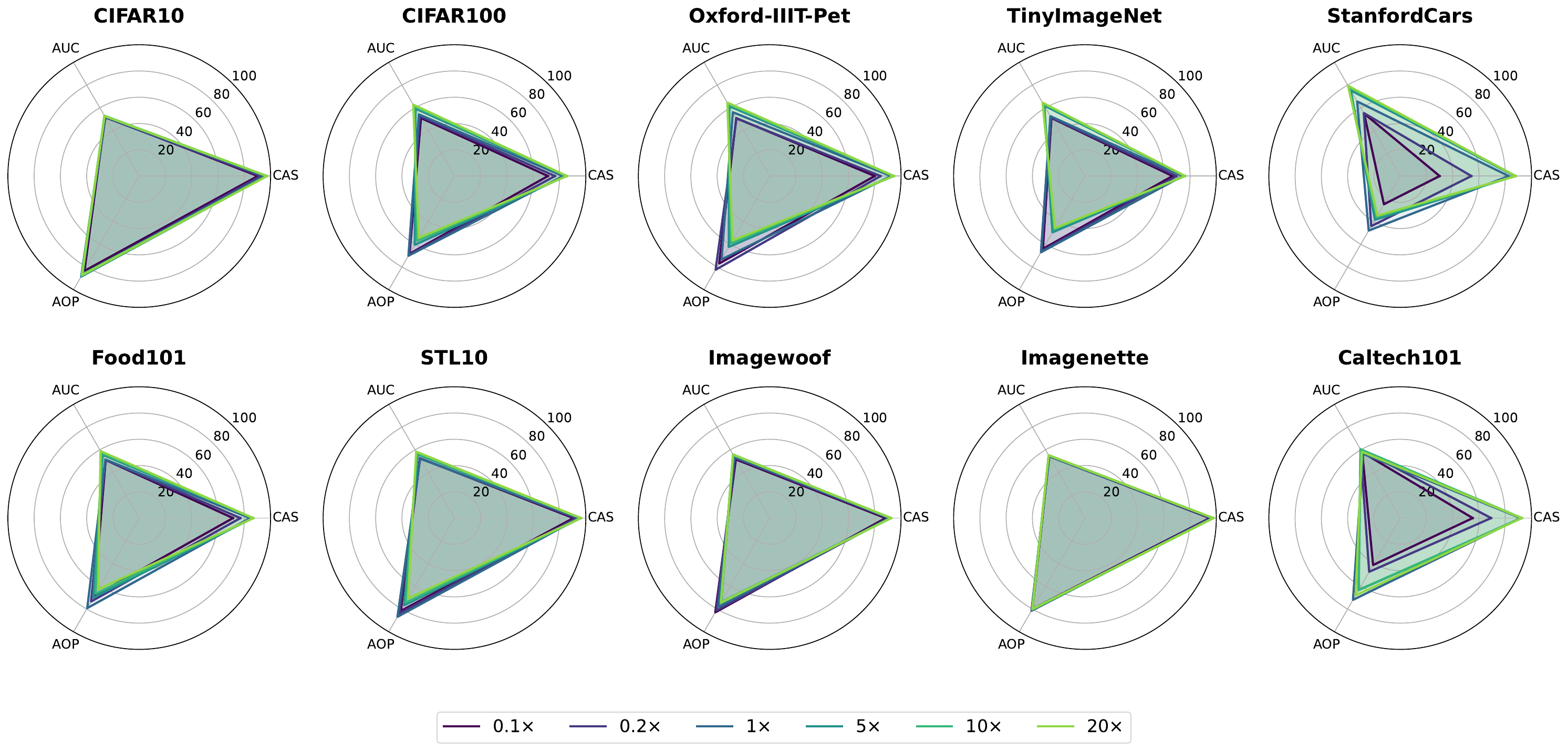}
    \caption{\textbf{Radar charts} comparing CAS, AUC$_{MIA}$, and AOP for all datasets at each synthetic dataset size. Each chart plots the raw values of the three metrics (higher CAS and AOP, and lower AUC$_{MIA}$, are better). Note that lower AUC$_{MIA}$ values (closer to 50) indicate stronger privacy protection.}
    \label{fig:metrics_balance_radar}
\end{figure*}

\noindent The AUC$_{MIA}$ values reveal clear patterns in privacy risk as synthetic data size grows. An ideal privacy-preserving model has AUC$_{MIA}$ $= 50$, indicating that an attacker's success is no better than random guessing. In the experiments, models trained on the smallest synthetic datasets (0.1$\times$ or 0.2$\times$) indeed achieve AUC$_{MIA}$ values closest to 50, reflecting strong inherent privacy. However, as the synthetic dataset size increases, the models tend to become more vulnerable to MIAs (higher AUC$_{MIA}$). For example, on StanfordCars the AUC$_{MIA}$ increases from 55.00 at 0.1$\times$ to 79.44 at 20$\times$, indicating a significant erosion of privacy at extreme cardinalities. Oxford-IIIT-Pet and TinyImageNet similarly show substantial rises in membership inference susceptibility as more synthetic data are used. This pattern is not universal: notably, CIFAR10 maintains nearly stable AUC$_{MIA}$ values across all scales (ranging only from 51.53 to 53.03), suggesting that some models preserve privacy well regardless of synthetic data volume.

\noindent Analysing the average AUC$_{MIA}$ changes between successive dataset sizes (\ref{fig:auc_vs_cardinality}, right) provides further insight. The most significant jump in privacy risk occurs when expanding the synthetic dataset from 1$\times$ to 5$\times$, where the mean AUC$_{MIA}$ increases by 4.30. In comparison, the increase is only +1.72 going from 0.2$\times$ to 1$\times$, +1.09 from 5$\times$ to 10$\times$, and a minimal +0.69 from 10$\times$ to 20$\times$. This progression suggests that the initial expansion beyond the original dataset size has the largest impact on privacy, while further enlarging an already large synthetic set yields progressively smaller privacy degradations.

\noindent The joint accuracy-privacy trade-off is next considered using the AOP metric (higher AOP indicates a better balance of high accuracy and low privacy risk). Interestingly, the results suggest that moderate synthetic dataset sizes yield the best trade-off between performance and privacy. In 8 out of 10 datasets, the peak AOP is achieved at either 0.2$\times$ or 1$\times$ cardinality. In other words, synthetic datasets of a standard order of magnitude (comparable in size to the real dataset) often offer the optimal balance. Beyond this point, adding more data tends to reduce AOP, meaning that the marginal accuracy gains are outweighed by increased privacy leakage. For instance, increasing from 1$\times$ to 5$\times$ causes a marked drop in AOP for many tasks (an average change of $-7.87$ across datasets). Smaller declines are observed for further increases (mean $-1.81$ from 5$\times$ to 10$\times$ and $-1.10$ from 10$\times$ to 20$\times$). Conversely, going from extremely small to moderate synthetic sets can improve AOP: for example, on average AOP rises by $+4.44$ when increasing from 0.1$\times$ to 0.2$\times$, and by $+2.58$ from 0.2$\times$ to 1$\times$. This reflects the fact that a minimum quantity of synthetic data is required to achieve good accuracy without overly compromising privacy, whereas excessive synthetic data eventually incurs diminishing returns in accuracy alongside greater privacy risk.

\noindent Task complexity also plays a role in the privacy-utility trade-off. Datasets with very fine-grained classes or with limited real training examples show the largest early gains in AOP when increasing synthetic data. For instance, StanfordCars sees its AOP jump from 24.82 at 0.1$\times$ to 47.97 at 1$\times$, and Caltech101 improves from 41.28 to 71.93 over the same range. These substantial gains indicate that for challenging tasks, using a moderate amount of synthetic data greatly improves the balance between accuracy and privacy.

\begin{figure*}[t]
    \centering
    \includegraphics[width=0.95\textwidth]{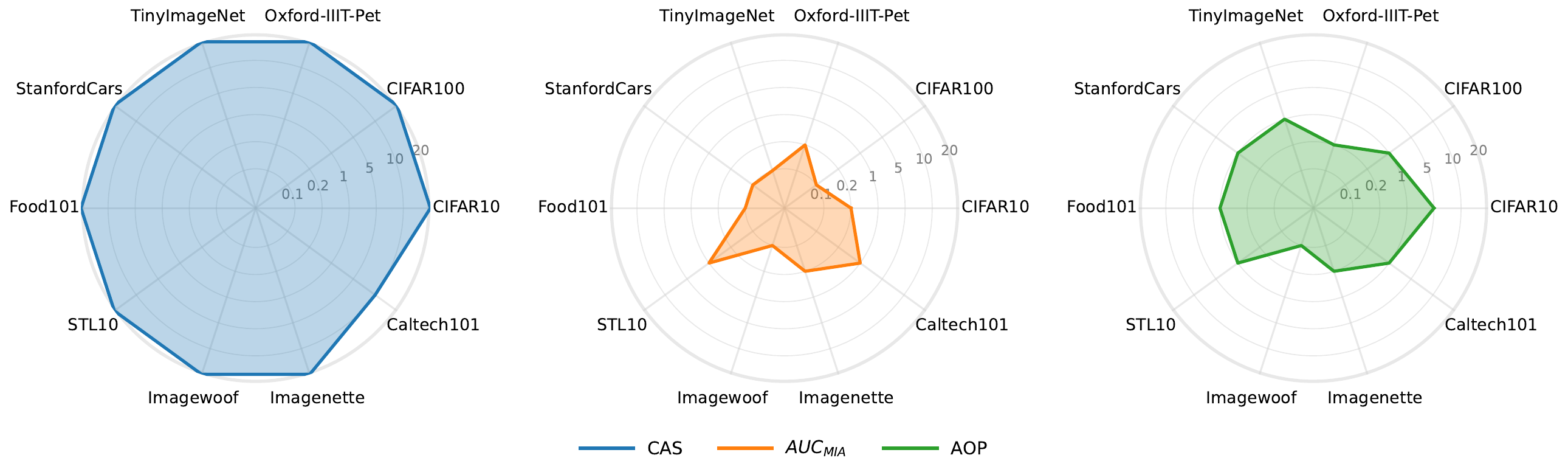}
    \caption{\textbf{“Best cardinality” spider plot} indicating which synthetic dataset size yields the optimal value for each metric on each dataset. Each spoke corresponds to a dataset, and markers on that spoke indicate the cardinality (0.1$\times$, 0.2$\times$, 1$\times$, 5$\times$, 10$\times$, or 20$\times$) at which CAS, AUC$_{MIA}$, or AOP is maximized (or in the case of AUC$_{MIA}$, minimised for best privacy).}
    \label{fig:metrics_best_cardinality}
\end{figure*}

\noindent Crucially, the privacy evaluation highlights a clear advantage of synthetic training data over real data. Across almost all settings, models trained on synthetic images provide better privacy protection (lower AUC$_{MIA}$ and higher AOP) than their counterparts trained on the original real datasets. With the exception of a few extreme cases – namely, the very lowest cardinalities for CIFAR10 and StanfordCars, and the two lowest for Caltech101, where the Students' accuracy was too low to yield a good trade-off – the synthetic data is always the more privacy-favourable choice. This is due to the inherent privacy properties it confers to the Student Classifiers (since synthetic images do not correspond to actual user data). \ref{fig:metrics_balance_radar} offers a consolidated view of CAS, AUC$_{MIA}$, and AOP across all datasets and cardinalities. Consistent with the earlier observations, it shows a clear inverse relationship between classification accuracy and privacy at high data volumes: as synthetic dataset size increases, accuracy improves but privacy risk (AUC$_{MIA}$) also rises, resulting in lower AOP. Nevertheless, because any size of synthetic dataset still yields significantly lower MIA vulnerability than using real data, one can choose an appropriate synthetic dataset size in TCKR to meet a desired accuracy target while still respecting a privacy threshold. As illustrated in \ref{fig:metrics_best_cardinality}, the optimal balance is typically achieved at a moderate scale (around 0.2$\times$-1$\times$ of the real dataset size). In summary, increasing synthetic data size consistently boosts model accuracy but gradually diminishes privacy; the TCKR pipeline provides the flexibility to navigate this trade-off by selecting a dataset size that maximises accuracy under acceptable privacy risk for the application at hand.

\begin{table*}[!t]
   \centering
   \setlength{\tabcolsep}{3pt}
   \caption{Comparison between Teacher and Student Classifiers in terms of Classification Accuracy, AUC$_{MIA}$, and AOP. \textbf{Bold} highlights the best score for each metric, while the performance gap ($\Delta$) between the Student and Teacher is reported at the bottom.}
   \footnotesize
   \begin{tabularx}{\textwidth}{l*{6}{Y}}
       \toprule
       & \multicolumn{2}{c}{Accuracy $\uparrow$} & \multicolumn{2}{c}{AUC$_{MIA}$ $\downarrow$} & \multicolumn{2}{c}{AOP $\uparrow$}\\
       \cmidrule(lr){2-3} \cmidrule(lr){4-5} \cmidrule(lr){6-7}
       Model & Teacher Classifier & Student Classifier & Teacher Classifier & Student Classifier & Teacher Classifier & Student Classifier \\
       \midrule
       CIFAR10         & \textbf{97.52} & 97.33 & 53.89 & \textbf{52.81} & 83.95 & \textbf{87.25}\\
       CIFAR100        & 85.49 & \textbf{85.74} & 70.32 & \textbf{62.80} & 43.22 & \textbf{54.35}\\
       Oxford-IIIT-Pet & 93.96 & \textbf{94.68} & 72.74 & \textbf{64.60} & 44.40 & \textbf{56.72}\\
       TinyImageNet    & 75.67 & \textbf{76.22} & 70.29 & \textbf{64.55} & 38.29 & \textbf{45.73}\\
       StanfordCars    & \textbf{88.22} & 88.00 & 82.53 & \textbf{79.44} & 32.38 & \textbf{34.86}\\
       Food101         & 86.79 & \textbf{86.93} & 65.63 & \textbf{58.90} & 50.37 & \textbf{62.64}\\
       STL10           & \textbf{96.74} & 96.36 & 65.89 & \textbf{58.38} & 55.71 & \textbf{70.68}\\
       Imagewoof       & \textbf{93.05} & 92.52 & 58.67 & \textbf{55.82} & 67.58 & \textbf{74.23}\\
       Imagenette      & \textbf{98.29} & 97.83 & 60.32 & \textbf{55.32} & 67.53 & \textbf{79.92}\\
       Caltech101      & 92.26 & \textbf{92.62} & 67.81 & \textbf{60.60} & 50.16 & \textbf{63.05}\\
       \midrule
       Min $\Delta$         & - & - 0.53 & - & - 1.08 & - & + 2.48 \\
       Mean $\Delta$        & - & + 0.02 & - & - 5.49 & - & + 9.58 \\
       Max $\Delta$         & - & + 0.72 & - & - 8.14 & - & + 14.97 \\
       \bottomrule
   \end{tabularx}
   \label{tab:student_vs_teacher}
\end{table*}

%


\subsection{Final Comparison}
\label{results:final_teacher_vs_student}

\noindent Finally, a comparison is made between the best-performing TCKR Student Classifiers and the original Teacher Classifiers trained on real data. \ref{tab:student_vs_teacher} summarises this comparison across three key metrics: Accuracy (CAS for Students vs. standard accuracy for Teachers), AUC$_{MIA}$, and AOP. For each dataset, the Student model that achieved the highest CAS (marked by a star in \ref{fig:cas_vs_cardinality}) is considered and its performance is contrasted with the corresponding Teacher.

\noindent The results reveal that the Student Classifiers consistently attain classification performance comparable to, and in some cases better than, the Teacher Classifiers. In half of the evaluated datasets, the Student actually outperforms its Teacher, achieving a higher CAS than the Teacher's accuracy on real data. These gains range from a modest +0.14 (on Food101) to a notable +0.72 (on Oxford-IIIT-Pet). This is a remarkable finding: a model trained exclusively on synthetic images can match or even slightly exceed the accuracy of a model trained on the original real dataset. One possible explanation is that the synthetic data generation helps to mitigate certain biases or limitations present in the real training data, while still preserving the essential class-specific information needed for effective classification.

\noindent In terms of privacy, the advantages of the TCKR approach are even more pronounced. Across all datasets, Student Classifiers exhibit substantially lower AUC$_{MIA}$ values than Teachers, indicating greater resistance to Membership Inference Attacks. The average AUC$_{MIA}$ reduction is 5.49 points, with the largest drop being 8.14 points (for Oxford-IIIT-Pet). In other words, models trained on TCKR-synthesised data are significantly less vulnerable to privacy attacks compared to those trained on real data. Some of the most pronounced privacy gains occur on datasets with rich visual diversity (e.g., Oxford-IIIT-Pet, TinyImageNet, Caltech101), suggesting that the text-conditioned generative approach is especially effective in complex domains for reducing memorisation of specific training examples.

\noindent The AOP metric further highlights the improved accuracy-privacy balance achieved by TCKR. On average, the Student Classifiers' AOP is higher by 9.58 points relative to their Teacher counterparts, with a maximum improvement of 14.97 (observed for Imagenette). These substantial AOP gains demonstrate that the TCKR pipeline yields models with a far better trade-off between utility and privacy: the Students maintain high accuracy while greatly lowering privacy risks, whereas the Teachers trained on real data have inferior trade-offs.

\noindent It is important to note that these privacy benefits come with virtually no cost to accuracy. The largest observed accuracy deficit for a Student relative to its Teacher in the experiments is only 0.53, and on average the difference in accuracy is essentially zero (the mean CAS difference is +0.02 in favour of the Students). In practice, this means that replacing real training data with TCKR synthetic data does not degrade classifier accuracy at all, and in many cases actually improves it, while yielding significant privacy advantages. This finding stands in stark contrast to the common assumption that models trained on synthetic data will perform worse on real-world evaluations. The results demonstrate that, with a carefully designed pipeline like TCKR, it is possible to achieve the opposite outcome: improved model performance alongside enhanced privacy.

\noindent In summary, the TCKR pipeline effectively recycles knowledge from pre-trained generative models to create synthetic datasets that rival real data in utility. The Student Classifiers trained on these synthetic sets achieve accuracy on par with (or above) those trained on real data, and are uniformly more resilient to Membership Inference Attacks. These findings provide strong empirical evidence that high-fidelity synthetic data can serve as a viable replacement for real images in training classifiers, enabling both high performance and improved privacy across a wide range of image recognition tasks.


\section{Conclusions}
\label{sec:conclusions}
\noindent This research presents Text-Conditioned Knowledge Recycling (TCKR) as a novel pipeline for creating high-quality synthetic training data for downstream image classification tasks. 
The experimental results demonstrate that the synthetic data generated through the TCKR pipeline not only serves as an adequate substitute for real training images but can actually enhance classifier performance in specific contexts. The combination of text-conditioned diffusion models, dynamic captioning, and Generative Knowledge Distillation techniques create synthetic datasets that capture essential visual features whilst introducing beneficial variations that might be absent in more limited real datasets.

\noindent A particularly significant finding from this investigation relates to the scaling properties of synthetic data. The relationship between synthetic dataset size and classification accuracy follows a consistent pattern: performance improves with increasing data volume, though with diminishing returns at extremely large scales. The peak accuracy is typically observed at the maximum tested dataset size (20$\times$ the original dataset length), suggesting untapped potential for further improvements with even larger synthetic datasets. This scaling behaviour opens the avenue to numerous solutions, addressing challenges where either data collection is constrained or manual annotation would require extensive human effort.

\noindent The privacy advantages of synthetic-data-trained models constitute another crucial dimension of this work. Through detailed analysis, this research identifies an optimal operating point where moderate synthetic dataset sizes (comparable to the original real dataset) offer the best balance between classification accuracy and privacy preservation. At this optimal point, membership inference risk remains minimal whilst classification performance equals or slightly exceeds that of real-data training. This finding is particularly relevant for applications where data privacy concerns are paramount.

\noindent The limitations of the current approach provide clear directions for future research. Whilst TCKR has proven effective for classification tasks, its extension to other computer vision problems such as object detection, segmentation, or instance recognition requires further investigation. The reliance on captions derived from original data also presents an opportunity for innovation through synthetic caption generation or language model integration. Additionally, the potential for recursive synthetic data generation~-- using models trained on synthetic data to generate subsequent training datasets~-- remains unexplored and could yield compounding benefits.

\noindent The TCKR methodology represents a significant advancement in developing privacy-preserving machine learning systems that do not sacrifice performance. By demonstrating that carefully constructed synthetic data can match or exceed the utility of real data whilst offering enhanced privacy characteristics, this work challenges the conventional assumption that synthetic training data necessarily leads to performance degradation. Instead, it provides evidence that thoughtfully designed synthetic data pipelines can simultaneously address multiple challenges in modern machine learning: data scarcity, privacy concerns, and annotation costs. Future research building upon these findings could establish synthetic data generation as a fundamental paradigm across diverse machine learning applications beyond computer vision.

\section{Acknowledgements}\label{sec:Acknowledgements}
\noindent This paper is supported by the FAIR (Future Artificial Intelligence Research) project, funded by the NextGenerationEU program within the PNRR-PE-AI scheme (M4C2, investment 1.3, line on Artificial Intelligence).

\bibliography{references}

\begin{thebibliography}{10}
\expandafter\ifx\csname url\endcsname\relax
  \def\url#1{\texttt{#1}}\fi
\expandafter\ifx\csname urlprefix\endcsname\relax\def\urlprefix{URL }\fi
\expandafter\ifx\csname href\endcsname\relax
  \def\href#1#2{#2} \def\path#1{#1}\fi

\bibitem{bansal2022systematic}
M.~A. Bansal, D.~R. Sharma, D.~M. Kathuria, A systematic review on data scarcity problem in deep learning: solution and applications, ACM Computing Surveys (Csur) 54~(10s) (2022) 1--29.

\bibitem{litjens2017survey}
G.~Litjens, T.~Kooi, B.~E. Bejnordi, A.~A.~A. Setio, F.~Ciompi, M.~Ghafoorian, J.~A. Van Der~Laak, B.~Van~Ginneken, C.~I. S{\'a}nchez, A survey on deep learning in medical image analysis, Medical image analysis 42 (2017) 60--88.

\bibitem{figueira2022survey}
A.~Figueira, B.~Vaz, Survey on synthetic data generation, evaluation methods and gans, Mathematics 10~(15) (2022) 2733.

\bibitem{rombach2022high}
R.~Rombach, A.~Blattmann, D.~Lorenz, P.~Esser, B.~Ommer, High-resolution image synthesis with latent diffusion models, in: Proceedings of the IEEE/CVF conference on computer vision and pattern recognition, 2022, pp. 10684--10695.

\bibitem{shokri2017membership}
R.~Shokri, M.~Stronati, C.~Song, V.~Shmatikov, Membership inference attacks against machine learning models, in: Symposium on Security and Privacy, 2017.

\bibitem{li2023blip}
J.~Li, D.~Li, S.~Savarese, S.~Hoi, Blip-2: Bootstrapping language-image pre-training with frozen image encoders and large language models, in: International conference on machine learning, PMLR, 2023, pp. 19730--19742.

\bibitem{lomurno2025synthetic}
E.~Lomurno, M.~Matteucci, Synthetic image learning: Preserving performance and preventing membership inference attacks, Pattern Recognition Letters (2025).

\bibitem{goodfellow2014generative}
I.~Goodfellow, J.~Pouget-Abadie, M.~Mirza, B.~Xu, D.~Warde-Farley, S.~Ozair, A.~Courville, Y.~Bengio, Generative adversarial networks, Advances in Neural Information Processing Systems (2014).

\bibitem{ho2020denoising}
J.~Ho, A.~Jain, P.~Abbeel, Denoising diffusion probabilistic models, Advances in neural information processing systems 33 (2020) 6840--6851.

\bibitem{mirza2014conditional}
M.~Mirza, Conditional generative adversarial nets, arXiv preprint arXiv:1411.1784 (2014).

\bibitem{brock2018large}
A.~Brock, Large scale gan training for high fidelity natural image synthesis, arXiv preprint arXiv:1809.11096 (2018).

\bibitem{ku2023textcontrolgan}
H.~Ku, M.~Lee, Textcontrolgan: Text-to-image synthesis with controllable generative adversarial networks, Applied Sciences 13~(8) (2023) 5098.

\bibitem{tao2023galip}
M.~Tao, B.-K. Bao, H.~Tang, C.~Xu, Galip: Generative adversarial clips for text-to-image synthesis, in: Proceedings of the IEEE/CVF Conference on Computer Vision and Pattern Recognition, 2023, pp. 14214--14223.

\bibitem{radford2021learning}
A.~Radford, J.~W. Kim, C.~Hallacy, A.~Ramesh, G.~Goh, S.~Agarwal, G.~Sastry, A.~Askell, P.~Mishkin, J.~Clark, et~al., Learning transferable visual models from natural language supervision, in: International conference on machine learning, PMLR, 2021, pp. 8748--8763.

\bibitem{nichol2021glide}
A.~Nichol, P.~Dhariwal, A.~Ramesh, P.~Shyam, P.~Mishkin, B.~McGrew, I.~Sutskever, M.~Chen, Glide: Towards photorealistic image generation and editing with text-guided diffusion models, arXiv preprint arXiv:2112.10741 (2021).

\bibitem{yang2024mastering}
L.~Yang, Z.~Yu, C.~Meng, M.~Xu, S.~Ermon, C.~Bin, Mastering text-to-image diffusion: Recaptioning, planning, and generating with multimodal llms, in: Forty-first International Conference on Machine Learning, 2024.

\bibitem{saharia2022photorealistic}
C.~Saharia, W.~Chan, S.~Saxena, L.~Li, J.~Whang, E.~L. Denton, K.~Ghasemipour, R.~Gontijo~Lopes, B.~Karagol~Ayan, T.~Salimans, et~al., Photorealistic text-to-image diffusion models with deep language understanding, Advances in neural information processing systems 35 (2022) 36479--36494.

\bibitem{betker2023improving}
J.~Betker, G.~Goh, L.~Jing, T.~Brooks, J.~Wang, L.~Li, L.~Ouyang, J.~Zhuang, J.~Lee, Y.~Guo, et~al., Improving image generation with better captions, Computer Science. https://cdn. openai. com/papers/dall-e-3. pdf 2~(3) (2023) 8.

\bibitem{ruiz2023dreambooth}
N.~Ruiz, Y.~Li, V.~Jampani, Y.~Pritch, M.~Rubinstein, K.~Aberman, Dreambooth: Fine tuning text-to-image diffusion models for subject-driven generation, in: Proceedings of the IEEE/CVF conference on computer vision and pattern recognition, 2023, pp. 22500--22510.

\bibitem{gal2022image}
R.~Gal, Y.~Alaluf, Y.~Atzmon, O.~Patashnik, A.~H. Bermano, G.~Chechik, D.~Cohen-Or, An image is worth one word: Personalizing text-to-image generation using textual inversion, arXiv preprint arXiv:2208.01618 (2022).

\bibitem{shumailov2024ai}
I.~Shumailov, Z.~Shumaylov, Y.~Zhao, N.~Papernot, R.~Anderson, Y.~Gal, Ai models collapse when trained on recursively generated data, Nature 631~(8022) (2024) 755--759.

\bibitem{sariyildiz2023fake}
M.~B. Sar{\i}y{\i}ld{\i}z, K.~Alahari, D.~Larlus, Y.~Kalantidis, Fake it till you make it: Learning transferable representations from synthetic imagenet clones, in: Proceedings of the IEEE/CVF Conference on Computer Vision and Pattern Recognition, 2023, pp. 8011--8021.

\bibitem{lomurno2024stable}
E.~Lomurno, M.~D'Oria, M.~Matteucci, et~al., Stable diffusion dataset generation for downstream classification tasks, in: European Symposium on Artificial Neural Networks, Computational Intelligence and Machine Learning, 2024, pp. N--A.

\bibitem{lampis2023bridging}
A.~Lampis, E.~Lomurno, M.~Matteucci, Bridging the gap: Enhancing the utility of synthetic data via post-processing techniques, British Machine Vision Conference (2023).

\bibitem{shipard2023diversity}
J.~Shipard, A.~Wiliem, K.~N. Thanh, W.~Xiang, C.~Fookes, Diversity is definitely needed: Improving model-agnostic zero-shot classification via stable diffusion, in: Proceedings of the IEEE/CVF Conference on Computer Vision and Pattern Recognition, 2023, pp. 769--778.

\bibitem{lei2023image}
S.~Lei, H.~Chen, S.~Zhang, B.~Zhao, D.~Tao, Image captions are natural prompts for text-to-image models, arXiv preprint arXiv:2307.08526 (2023).

\bibitem{li2024semantic}
B.~Li, X.~Xu, X.~Wang, Y.~Hou, Y.~Feng, F.~Wang, X.~Zhang, Q.~Zhu, W.~Che, Semantic-guided generative image augmentation method with diffusion models for image classification, in: Proceedings of the AAAI Conference on Artificial Intelligence, Vol.~38, 2024, pp. 3018--3027.

\bibitem{lomurno2025federated}
E.~Lomurno, M.~Matteucci, Federated knowledge recycling: Privacy-preserving synthetic data sharing, Pattern Recognition Letters (2025).

\bibitem{fredrikson2015model}
M.~Fredrikson, S.~Jha, T.~Ristenpart, Model inversion attacks that exploit confidence information and basic countermeasures, in: Proceedings of the ACM SIGSAC Conference on Computer and Communications Security, 2015.

\bibitem{tramer2016stealing}
F.~Tram{\`e}r, F.~Zhang, A.~Juels, M.~K. Reiter, T.~Ristenpart, Stealing machine learning models via prediction $\{$APIs$\}$, in: USENIX Security Symposium, 2016.

\bibitem{biggio2012poisoning}
B.~Biggio, B.~Nelson, P.~Laskov, Poisoning attacks against support vector machines, in: Proceedings of the International Conference on Machine Learning, 2012.

\bibitem{carlini2022membership}
N.~Carlini, S.~Chien, M.~Nasr, S.~Song, A.~Terzis, F.~Tramer, Membership inference attacks from first principles, in: 2022 IEEE Symposium on Security and Privacy (SP), IEEE, 2022, pp. 1897--1914.

\bibitem{abadi2016deep}
M.~Abadi, A.~Chu, I.~Goodfellow, H.~B. McMahan, I.~Mironov, K.~Talwar, L.~Zhang, Deep learning with differential privacy, in: Proceedings of the ACM SIGSAC Conference on Computer and Communications Security, 2016.

\bibitem{lomurno2022utility}
E.~Lomurno, M.~Matteucci, On the utility and protection of optimization with differential privacy and classic regularization techniques, in: International Conference on Machine Learning, Optimization, and Data Science, 2022.

\bibitem{hu2022defending}
L.~Hu, J.~Li, G.~Lin, S.~Peng, Z.~Zhang, Y.~Zhang, C.~Dong, Defending against membership inference attacks with high utility by gan, IEEE Transactions on Dependable and Secure Computing 20~(3) (2022) 2144--2157.

\bibitem{schuhmann2022laion}
C.~Schuhmann, R.~Beaumont, R.~Vencu, C.~Gordon, R.~Wightman, M.~Cherti, T.~Coombes, A.~Katta, C.~Mullis, M.~Wortsman, et~al., Laion-5b: An open large-scale dataset for training next generation image-text models, Advances in Neural Information Processing Systems 35 (2022) 25278--25294.

\bibitem{hu2021lora}
E.~J. Hu, P.~Wallis, Z.~Allen-Zhu, Y.~Li, S.~Wang, L.~Wang, W.~Chen, et~al., Lora: Low-rank adaptation of large language models, in: International Conference on Learning Representations, 2022.

\bibitem{miller1995wordnet}
G.~A. Miller, Wordnet: a lexical database for english, Communications of the ACM 38~(11) (1995) 39--41.

\bibitem{zhang2022opt}
S.~Zhang, S.~Roller, N.~Goyal, M.~Artetxe, M.~Chen, S.~Chen, C.~Dewan, M.~Diab, X.~Li, X.~V. Lin, et~al., Opt: Open pre-trained transformer language models, arXiv preprint arXiv:2205.01068 (2022).

\bibitem{wolf2020transformers}
T.~Wolf, L.~Debut, V.~Sanh, J.~Chaumond, C.~Delangue, A.~Moi, P.~Cistac, T.~Rault, R.~Louf, M.~Funtowicz, et~al., Transformers: State-of-the-art natural language processing, in: Proceedings of the 2020 conference on empirical methods in natural language processing: system demonstrations, 2020, pp. 38--45.

\bibitem{ravuri2019classification}
S.~Ravuri, O.~Vinyals, Classification accuracy score for conditional generative models, Advances in neural information processing systems 32 (2019).

\bibitem{lomurno2024pomonag}
E.~Lomurno, S.~Mariani, M.~Monti, M.~Matteucci, Pomonag: Pareto-optimal many-objective neural architecture generator, arXiv preprint arXiv:2409.20447 (2024).

\bibitem{deng2009imagenet}
J.~Deng, W.~Dong, R.~Socher, L.-J. Li, K.~Li, L.~Fei-Fei, Imagenet: A large-scale hierarchical image database, in: 2009 IEEE conference on computer vision and pattern recognition, Ieee, 2009, pp. 248--255.

\bibitem{howard2019searching}
A.~Howard, M.~Sandler, G.~Chu, L.-C. Chen, B.~Chen, M.~Tan, W.~Wang, Y.~Zhu, R.~Pang, V.~Vasudevan, et~al., Searching for mobilenetv3, in: Proceedings of the IEEE/CVF international conference on computer vision, 2019, pp. 1314--1324.

\bibitem{krizhevsky2009learning}
A.~Krizhevsky, G.~Hinton, et~al., Learning multiple layers of features from tiny images (2009).

\bibitem{parkhi2012cats}
O.~M. Parkhi, A.~Vedaldi, A.~Zisserman, C.~Jawahar, Cats and dogs, in: 2012 IEEE conference on computer vision and pattern recognition, IEEE, 2012, pp. 3498--3505.

\bibitem{le2015tiny}
Y.~Le, X.~Yang, Tiny imagenet visual recognition challenge, CS 231N 7~(7) (2015) 3.

\bibitem{krause20133d}
J.~Krause, M.~Stark, J.~Deng, L.~Fei-Fei, 3d object representations for fine-grained categorization, in: 2013 IEEE international conference on computer vision workshops, IEEE, 2013, pp. 554--561.

\bibitem{bossard2014food}
L.~Bossard, M.~Guillaumin, L.~Van~Gool, Food-101--mining discriminative components with random forests, in: Computer vision--ECCV 2014: 13th European conference, zurich, Switzerland, September 6-12, 2014, proceedings, part VI 13, Springer, 2014, pp. 446--461.

\bibitem{coates2011analysis}
A.~Coates, A.~Ng, H.~Lee, An analysis of single-layer networks in unsupervised feature learning, in: Proceedings of the fourteenth international conference on artificial intelligence and statistics, JMLR Workshop and Conference Proceedings, 2011, pp. 215--223.

\bibitem{howard2020fastai}
J.~Howard, S.~Gugger, Fastai: a layered api for deep learning, Information 11~(2) (2020) 108.

\bibitem{fei2004learning}
L.~Fei-Fei, R.~Fergus, P.~Perona, Learning generative visual models from few training examples: An incremental bayesian approach tested on 101 object categories, in: 2004 conference on computer vision and pattern recognition workshop, IEEE, 2004, pp. 178--178.

\bibitem{lomurno2023discriminative}
E.~Lomurno, A.~Archetti, F.~Ausonio, M.~Matteucci, et~al., Discriminative adversarial privacy: balancing accuracy and membership privacy in neural networks, in: The 34th British Machine Vision Conference Proceedings, BMVA, 2023, pp. N--A.

\bibitem{keys1981cubic}
R.~Keys, Cubic convolution interpolation for digital image processing, IEEE transactions on acoustics, speech, and signal processing 29~(6) (1981) 1153--1160.

\bibitem{han2013comparison}
D.~Han, Comparison of commonly used image interpolation methods, in: Conference of the 2nd International Conference on Computer Science and Electronics Engineering (ICCSEE 2013), Atlantis Press, 2013, pp. 1556--1559.

\bibitem{patel2013review}
V.~Patel, K.~Mistree, A review on different image interpolation techniques for image enhancement, International Journal of Emerging Technology and Advanced Engineering 3~(12) (2013) 129--133.

\bibitem{anthropic2024claude}
Anthropic, Claude (version 3.5 sonnet), https://www.anthropic.com/claude/sonnet (2024).

\bibitem{loshchilov2017decoupled}
I.~Loshchilov, Decoupled weight decay regularization, arXiv preprint arXiv:1711.05101 (2017).

\bibitem{cai2019once}
H.~Cai, C.~Gan, T.~Wang, Z.~Zhang, S.~Han, Once-for-all: Train one network and specialize it for efficient deployment, arXiv preprint arXiv:1908.09791 (2019).

\bibitem{hendrycks2019augmix}
D.~Hendrycks, N.~Mu, E.~D. Cubuk, B.~Zoph, J.~Gilmer, B.~Lakshminarayanan, Augmix: A simple data processing method to improve robustness and uncertainty, arXiv preprint arXiv:1912.02781 (2019).

\bibitem{szegedy2016rethinking}
C.~Szegedy, V.~Vanhoucke, S.~Ioffe, J.~Shlens, Z.~Wojna, Rethinking the inception architecture for computer vision, in: Proceedings of the IEEE conference on computer vision and pattern recognition, 2016, pp. 2818--2826.

\bibitem{zhang2017mixup}
H.~Zhang, mixup: Beyond empirical risk minimization, arXiv preprint arXiv:1710.09412 (2017).

\end{thebibliography}
\bibliographystyle{elsarticle-num}

\newpage
\clearpage

\section*{Appendix A: Generator Selection and Fine-Tuning}
This appendix provides supporting experiments for the Generator in Text-Conditioned Knowledge Recycling (TCKR), divided into model selection and fine-tuning strategy analyses. Each part underlines methodological differences observed during development.

\begin{table}[t]
    \centering
    \setlength{\tabcolsep}{3pt}
    \caption{CAS results for different Stable Diffusion versions (fine-tuned with LoRA~\cite{hu2021lora} on full data, class-only prompt). Best per dataset in \textbf{bold}.}
    \footnotesize
    \begin{tabularx}{\textwidth}{l*{3}{Y}}
        \toprule
        \textbf{Dataset} & \multicolumn{3}{c}{\textbf{CAS with \textit{Generator}:}} \\
        \cmidrule(lr){2-4}
         & SD 1.5 & SD 2.0 & SDXL\\ 
        \midrule
        CIFAR10         & 16.69 & \textbf{30.35} & 24.20\\
        CIFAR100        & 7.21 & \textbf{7.66} & 3.90\\
        Oxford-IIIT-Pet & 1.89 & \textbf{3.25} & 2.79\\
        \bottomrule
    \end{tabularx}
    \label{tab:generator_selection}
\end{table}

\subsection*{A.1~~Generator Model Comparison}
\label{sec:gen_model_comp}
\begin{figure}[t]
    \centering
    \includegraphics[width=\textwidth]{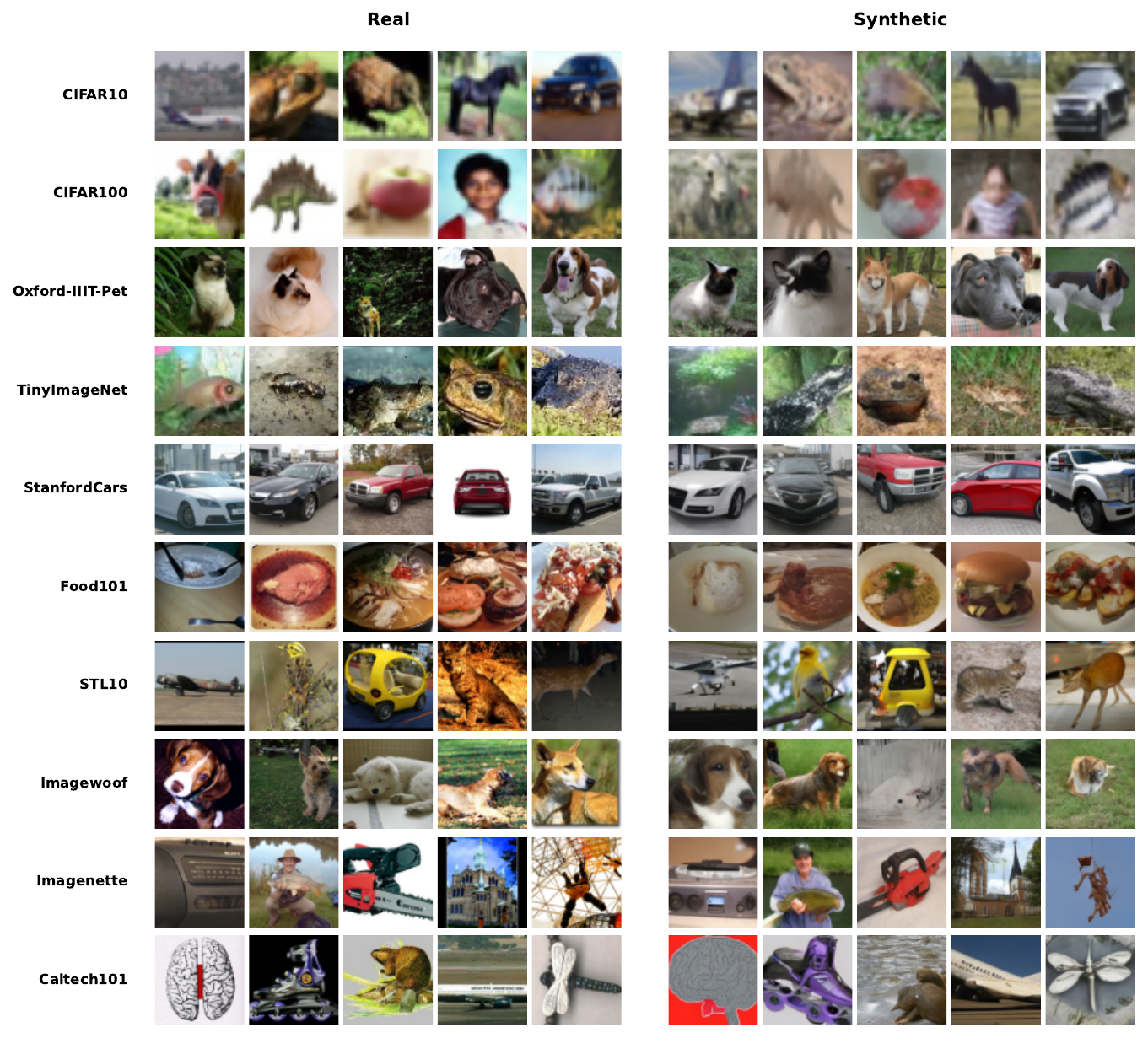}
    \caption{Qualitative comparison between real and synthetic images (generated after LoRA adaptation). For each dataset, the first five samples from different classes are shown.}
    \label{fig:real_vs_synthetic_images}
\end{figure}

To determine a suitable image Generator, Stable Diffusion models 1.5, 2.0, and SDXL~\cite{ho2020denoising} are compared on downstream Classification Accuracy Score (CAS)~\cite{ravuri2019classification}. Each model is fine-tuned on the entire real dataset (128$\times$128 resolution) with textual conditioning consisting only of the class corresponding to each image, then used to synthesise a dataset (0.1$\times$ cardinality, 224$\times$224 resolution) with prompts of form \textit{``n: d''} (class name + fixed description from Claude 3.5 Sonnet~\cite{anthropic2024claude}). Table~\ref{tab:generator_selection} reports CAS on CIFAR10, CIFAR100, and Oxford-IIIT-Pet. Stable Diffusion 2.0 outperforms SD1.5 and SDXL across all datasets, despite SDXL's theoretical advantages (higher output fidelity, better prompt understanding). This suggests that under a constrained fine-tuning regime (LoRA, 3 epochs), Stable Diffusion 2.0 yields more useful synthetic data. Figure~\ref{fig:real_vs_synthetic_images} shows real vs. Stable Diffusion 2.0 synthetic images, illustrating Stable Diffusion 2.0's high-fidelity generation of diverse class features.

\subsection*{A.2~~Optimising Generator Fine-Tuning}
\label{sec:gen_finetune_opt}
Next, Stable Diffusion 2.0's fine-tuning configuration is optimised. The testing includes:
\begin{itemize}
    \item \textbf{Prompt format:} class name only (\textit{``n''}) vs. class + description (\textit{``n: d''}).
    \item \textbf{Fine-tuning resolution:} $256\times256$ vs. $224\times224$ (the latter aligns with classifier input size).
    \item \textbf{Fine-tuning data:} 10\% of real training data (fixed for this analysis).
    \item \textbf{No fine-tuning baseline:} zero-shot Stable Diffusion 2.0 generation for reference.
\end{itemize}

Table~\ref{tab:finetuning_resolution} summarises CAS under each setting. For CIFAR10 and CIFAR100, fine-tuning with descriptive prompts at $224\times224$ yields the highest CAS, likely due to matching the classifier's native resolution and richer generator conditioning. Notably, using no fine-tuning (i.e., Stable Diffusion 2.0's pre-trained weights) is suboptimal for these datasets. Oxford-IIIT-Pet shows an anomaly: the zero-shot generator slightly exceeds fine-tuned variants. This may be because one-hot hard labels don't capture subtle breed similarities, causing fine-tuning to overfit spurious features; in contrast, Stable Diffusion 2.0's prior may generalise better given limited data. Nonetheless, the overall evidence favours fine-tuning at $224\times224$ with \textit{``n: d''}.

\begin{table*}[t]
    \centering
    \setlength{\tabcolsep}{3pt}
    \caption{CAS obtained when fine-tuning (or not) Stable Diffusion 2.0 with different prompt formats and image resolutions. \textbf{Bold} indicates the best performance for each dataset.}
    \footnotesize
    \begin{tabularx}{\textwidth}{l*{4}{Y}}
        \toprule
        \textbf{Dataset} & \multicolumn{4}{c}{\textbf{CAS with Stable Diffusion 2.0 Fine-Tuning \textit{Prompt $\circ$ Resolution:}}} \\
        \cmidrule(lr){2-5}
        & No Fine-Tuning & "$n$" $\circ$ 256$\times$256 & "$n: d$" $\circ$ 256$\times$256 & "$n: d$" $\circ$ 224$\times$224 \\
        \midrule
        CIFAR10         & 27.75 & 77.91 & 79.85 & \textbf{80.43}\\
        CIFAR100        & 18.18 & 43.03 & 43.92 & \textbf{45.43}\\
        Oxford-IIIT-Pet & \textbf{4.33} & 2.89 & 3.16 & 3.79\\
        \bottomrule
    \end{tabularx}
    \label{tab:finetuning_resolution}
\end{table*}

\section*{Appendix B: LoRA Adaptation Studies}
This appendix covers additional details of the LoRA~\cite{hu2021lora} adaptation technique, including hyperparameter choices and an ablation study on which parts of Stable Diffusion 2.0 to fine-tune.

\subsection*{B.1~~LoRA Fine-Tuning Setup}
\label{sec:lora_setup}
Stable Diffusion 2.0 is fine-tuned on each dataset using LoRA to inject new knowledge with minimal weight updates. Table~\ref{tab:lora} lists the involved LoRA hyperparameters. A low rank (4) is used for LoRA layers to limit added parameters, and only 3 epochs of fine-tuning are performed to preserve the model's generality. Training is in mixed precision to handle small batch size (1) for memory reasons. Data augmentations like random flips and center crops are enabled to expose the generator to varied views. LoRA is applied to all attention layers of the U-Net (query, key, value, output), but not to the variational autoencoder (VAE) for stability.

\begin{table*}[h] 
    \centering 
    \setlength{\tabcolsep}{3pt} 
    \caption{LoRA hyperparameters used for fine-tuning Stable Diffusion 2.0. The configuration includes training settings, model architecture adjustments, and attention mechanisms optimised for efficient adaptation whilst preserving generalisation capabilities.} 
    \footnotesize 
    \begin{tabularx}{\textwidth}{lY|lY} 
        \toprule 
        Hyperparameter & Value & Hyperparameter & Value\\
        \midrule
        Adaptation Epochs & 3 & Precision & Mixed\\ 
        Optimiser & AdamW~\cite{loshchilov2017decoupled} & Random Flip & True\\ 
        Learning Rate & 1$\times$10$^{-4}$ & Center Crop & True\\ 
        Learning Rate Scheduler & Constant & Adapt Query Attention & True\\ 
        Batch Size & 1 & Adapt Key Attention & True\\ 
        Rank & 4 & Adapt Value Attention & True\\ 
        Resolution & 224$\times$224 & Adapt Output Attention & True\\ 
        Gradient Accumulation & 4 & Adapt VAE Attention & False\\ 
        \bottomrule 
    \end{tabularx} 
    \label{tab:lora} 
\end{table*}

\begin{table*}[t]
    \centering
    \setlength{\tabcolsep}{3pt}
    \caption{Comparison of the CAS on CIFAR100 when adapting with LoRA only the U-Net versus both U-Net and text encoder (TE) of Stable Diffusion 2.0, evaluated for two different prompt formats: "$n: d$" and "$c$". \textbf{Bold} indicates the highest CAS in each row.}
    \footnotesize
    \begin{tabularx}{\textwidth}{lc*{10}{Y}}
        \toprule
        \textbf{LoRA Target} & \textbf{Prompt} & \multicolumn{10}{c}{\textbf{CAS with Fine-Tuning \textit{Samples Ratio}:}} \\
        \cmidrule(lr){3-12}
        & & $\frac{1}{10}$ & $\frac{2}{10}$ & $\frac{3}{10}$ & $\frac{4}{10}$ & $\frac{5}{10}$ & $\frac{6}{10}$ & $\frac{7}{10}$ & $\frac{8}{10}$ & $\frac{9}{10}$ & $\frac{10}{10}$ \\ 
        \midrule
        \multirow{2}{*}{U-Net} 
        & "$n: d$" & 45.43 & 40.74 & 46.09 & 47.88 & 46.87 & 46.69 & 44.50 & \textbf{48.09} & 46.68 & 47.37\\
        & "$c$" & 45.35 & 44.36 & 46.71 & \textbf{48.45} & 46.42 & 46.74 & 46.79 & 47.16 & 48.39 & 47.35\\
        \midrule
        \multirow{2}{*}{U-Net + TE}
        & "$n: d$" & \textbf{39.33} & 30.36 & 22.17 & 35.36 & 37.28 & 38.16 & \textbf{39.33} & 35.29 & 38.25 & 35.47\\
        & "$c$" & 34.31 & 30.37 & 38.70 & 41.59 & \textbf{42.86} & 38.20 & 41.30 & 39.21 & 38.72 & 40.47\\
        \bottomrule
    \end{tabularx}
    \label{tab:lora_unet_vs_te}
\end{table*}

\subsection*{B.2~~Ablation: U-Net vs. Text Encoder Tuning}
\label{sec:lora_ablation}
A further examination is conducted on whether extending LoRA to the Stable Diffusion 2.0 Text Encoder (TE) benefits performance. On CIFAR100, two LoRA targets are compared:
\begin{itemize}
    \item \textbf{U-Net:} LoRA applied to U-Net (image generation network) weights.
    \item \textbf{U-Net + TE:} LoRA applied to both U-Net and CLIP text encoder weights.
\end{itemize}
Each case was tested with two prompt regimes: fixed description \textit{``n: d''} vs. BLIP-2~\cite{li2023blip} caption \textit{``c''}. Table~\ref{tab:lora_unet_vs_te} shows CAS for each combination across varying fractions of real fine-tuning data (10\% up to 100\%). The trend is clear: focusing LoRA on U-Net yields higher CAS consistently. For example, at 100\% fine-tuning data, U-Net only (\textit{``c''} prompt) reaches 47.35 CAS vs. 40.47 for U-Net + TE. Even at lower data ratios, U-Net only configurations dominate. This is attributed to the Text Encoder already being pretrained on extensive language-image data; a lightweight LoRA update may disrupt its semantic alignment, whereas U-Net adaptation alone suffices to inject new visual details. Hence, all final experiments restrict LoRA to the U-Net. Additionally, within each LoRA target setting, using BLIP-2 captions (\textit{``c''}) tends to slightly improve CAS over fixed descriptions (\textit{``n: d''}) -- consistent with earlier prompt findings.

\begin{table*}[t]
    \centering
    \setlength{\tabcolsep}{3pt}
    \caption{Hyperparameters for training both the Teacher and the Student classifiers. The configuration includes model selection and initialisation, training schedule, and augmentation techniques applied during classifier training.}
    \footnotesize
    \begin{tabularx}{\textwidth}{lY|lY}
        \toprule
          Hyperparameter & Value & Hyperparameter & Value\\ 
          \midrule
          Search Space & MobileNetV3~\cite{howard2019searching} & Neural Architecture Search Method & POMONAG~\cite{lomurno2024pomonag}\\
          Pre-Training & OFA~\cite{cai2019once} (ImageNet-1k~\cite{deng2009imagenet}) & Input Dimension (Resizing) & 224$\times$224\\
          Epochs & 50 & Mixed Precision & True\\
          Batch Size & 96 & Major Augmentation & AugMix~\cite{hendrycks2019augmix}\\
          Early Stopping Patience & 30 & Label Smoothing~\cite{szegedy2016rethinking} & 0.1\\
          Optimiser & AdamW~\cite{loshchilov2017decoupled} & Random Horizontal Flip & 0.5\\
          Learning Rate & 1$\times$10$^{-3}$ & Padding (Constant) & 21\\
          Learning Rate Scheduler & Cosine Annealing & Random Crop & 224$\times$224\\
          Weight Decay & 5$\times$10$^{-5}$ & Mixup~\cite{zhang2017mixup} & 0.2\\
        \bottomrule
    \end{tabularx}
    \label{tab:classifier}
\end{table*}

\begin{figure}[t]
    \centering
    \includegraphics[width=\textwidth]{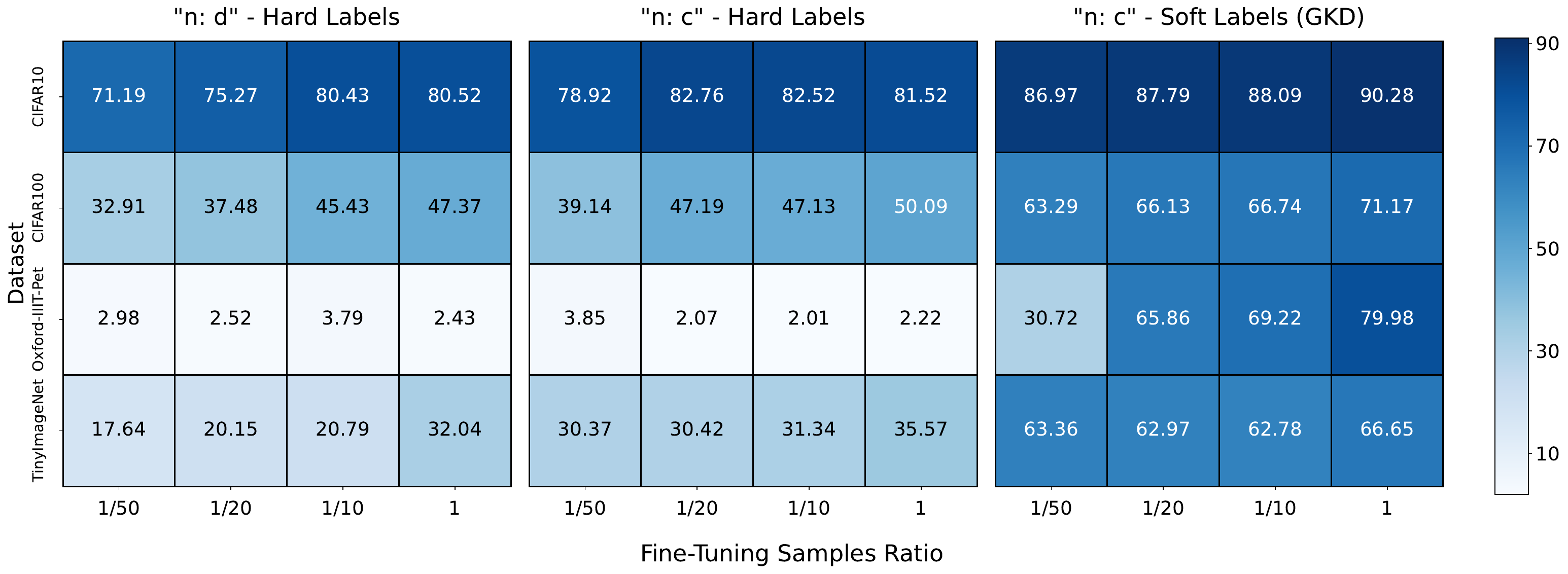}
    \caption{Heatmap depicting CAS values for three distinct \textit{prompt -- label} configurations across four datasets.}
    \label{fig:impact_blip2_gkd_heatmap}
\end{figure}

\begin{figure}[t]
    \centering
    \includegraphics[width=\textwidth]{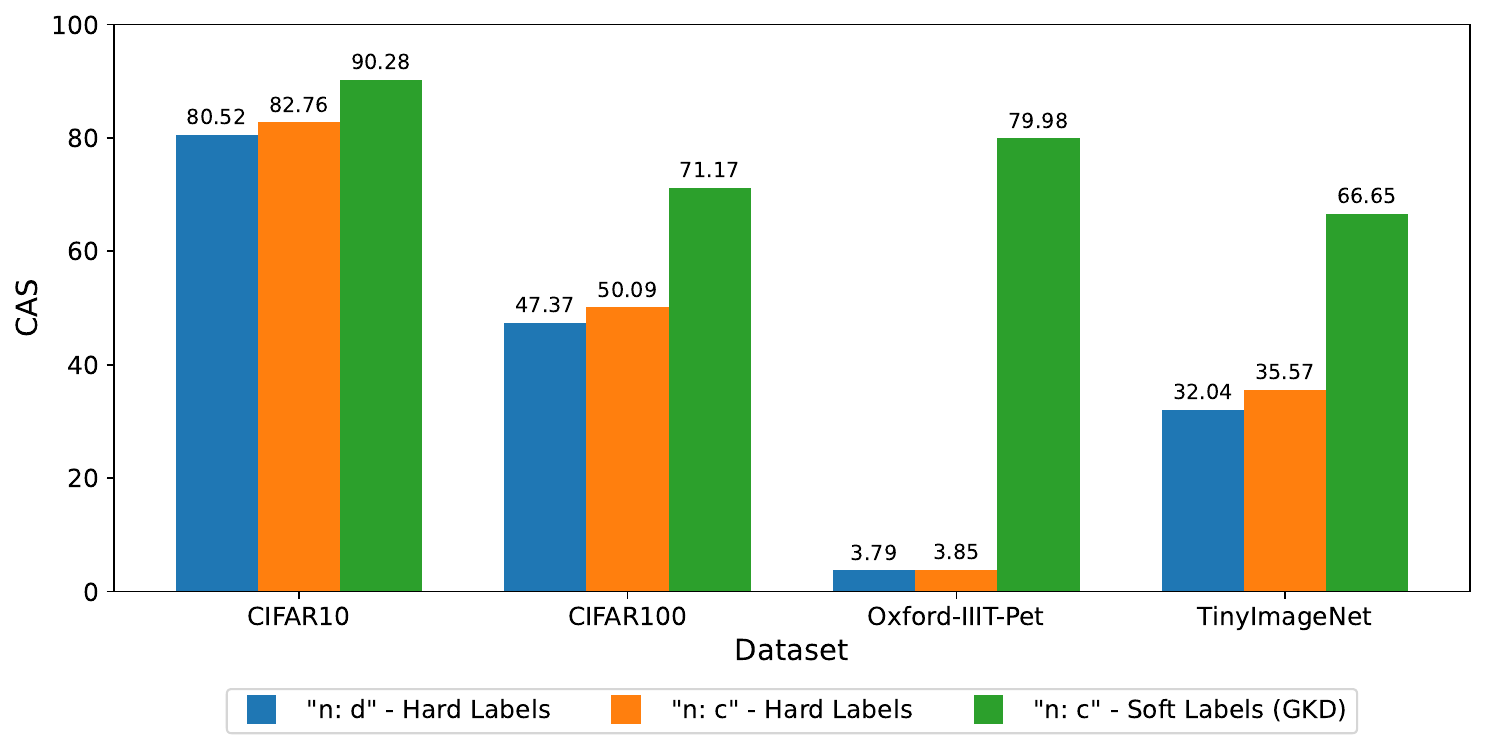}
    \caption{Comparison of the best CAS obtained in each configuration of Figure~\ref{fig:impact_blip2_gkd_heatmap}.}
    \label{fig:impact_blip2_gkd_chart}
\end{figure}

\begin{figure}[t]
    \centering
    \includegraphics[width=\textwidth]{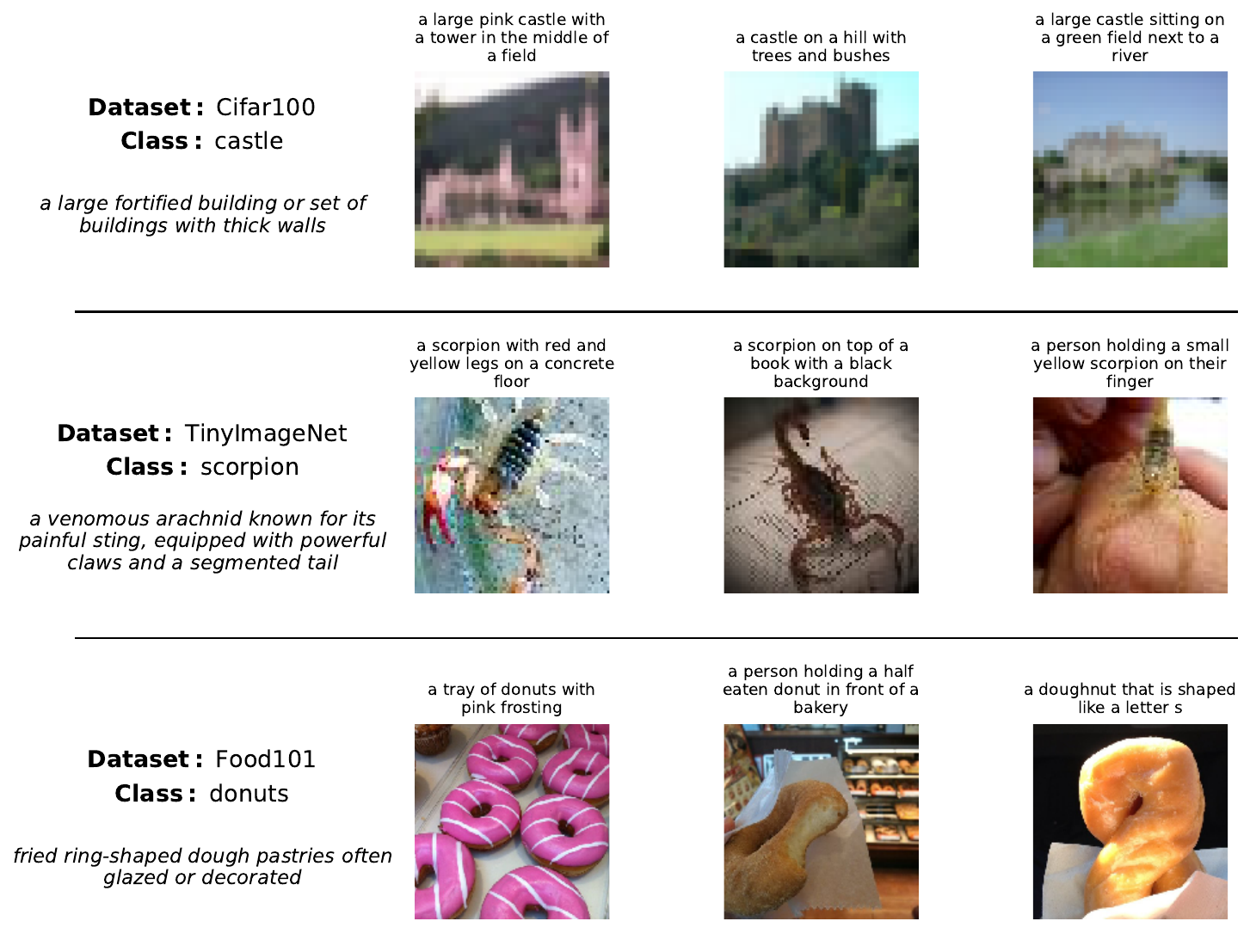}
    \caption{Qualitative comparison between fixed class descriptions and dynamic image-specific captions across datasets of varying resolutions. For each class, the fixed description (in \textit{italics}) was generated one time using Claude 3.5 Sonnet~\cite{anthropic2024claude}, providing a general class definition. In contrast, the image-specific captions above each real sample were dynamically generated by BLIP-2~\cite{li2023blip}, demonstrating its ability to accurately describe the class subject across different contexts, poses, and image qualities.}
    \label{fig:claude_vs_blip2_prompts}
\end{figure}

\section*{Appendix C: Classifier Training and Prompt/Label Analysis}
This appendix provides additional information on the Student Classifier training setup and experiments analysing the effects of prompts and label types on performance.
\subsection*{C.1~~Classifier Architecture and Hyperparameters}
\label{sec:clf_details}
A MobileNetV3-based classifier \cite{howard2019searching} is used, chosen via the multi-objective NAS method POMONAG~\cite{lomurno2024pomonag}. Table~\ref{tab:classifier} details the training configuration. The classifier is initialised from an OFA checkpoint~\cite{cai2019once} (pre-trained on ImageNet-1k~\cite{deng2009imagenet}) to leverage transferrable features. Training lasts up to 50 epochs with early stopping (patience 30). A standard image size of 224, batch size 96, and mixed precision are used. Data augmentation is crucial: AugMix~\cite{hendrycks2019augmix} is applied for robustness and Mixup \cite{zhang2017mixup} ($\alpha=0.2$) to mitigate overfitting, along with horizontal flips and random crops. Regularisation includes weight decay ($5\times10^{-5}$) and label smoothing (0.1) to improve generalisation. Optimisation is done with AdamW~\cite{loshchilov2017decoupled}, and the learning rate (1e-3) follows a cosine annealing schedule. These hyperparameters mirror common practice and were kept constant across all experiments for fairness.

\subsection*{C.2~~Prompt Format vs. Label Type: Extended Results}
\label{sec:prompt_label_analysis}
A 3-factor experiment is performed to disentangle the influence of (i) Generator fine-tuning data amount, (ii) prompt format, and (iii) label type on the final CAS. The factors are:
\begin{itemize}
    \item \textbf{Fine-tuning data ratio:} $\{1/50, 1/20, 1/10, 1\}$ of the real dataset used to fine-tune Stable Diffusion 2.0.
    \item \textbf{Prompt format:} \textit{``n: d''} fixed class description vs. \textit{``n: c''} dynamic BLIP-2 caption (both include class name).
    \item \textbf{Label type:} Hard labels (one-hot) vs. Soft labels via Generative Knowledge Distillation (GKD).
\end{itemize}

The synthetic dataset size is fixed at 0.1$\times$ per class. Figure~\ref{fig:impact_blip2_gkd_heatmap} visualises CAS outcomes for each combination on four representative datasets. Figure~\ref{fig:impact_blip2_gkd_chart} extracts the peak CAS per configuration. It is possible to underline the following findings:
\begin{itemize}
    \item Using more real data for generator fine-tuning monotonically improves CAS (most evident from left to right in each heatmap group). Especially for TinyImageNet and CIFAR100, low fine-tuning fractions significantly hurt performance.
    \item Switching from fixed to dynamic prompts yields consistent CAS gains under hard labels (compare first and second heatmap). This aligns with the earlier observation that BLIP-2 captions diversify generator inputs effectively.
    \item The largest jump comes from using soft labels: the third bar in Figure~\ref{fig:impact_blip2_gkd_chart} (``n: c'' - Soft Labels (GKD)) shows dramatically higher CAS than the second bar (``n: c'' - Hard Labels) in nearly every case. Oxford-IIIT-Pet is a striking example (soft labels boosting CAS from $\sim$3 to $\sim$80). Soft labels, by providing nuanced class similarity information, prevent the classifier from being misled by rigid one-hot targets in fine-grained scenarios.
\end{itemize}

To illustrate why dynamic prompts help, Figure~\ref{fig:claude_vs_blip2_prompts} shows examples for a class from CIFAR100 (\textit{castle}) and TinyImageNet (\textit{scorpion}), plus Food101 (\textit{donuts}). Each class has a single Claude 3.5 description (italic text) vs. multiple BLIP-2 captions above real images. The BLIP-2 captions capture specific details (e.g. a pink castle with a central tower, a scorpion on a person’s finger) that a generic description cannot, underscoring how \textit{``n: c''} prompts enrich training.

In summary, this analysis strongly supports three design choices: fine-tuning the generator on as much real data as possible, using dynamic image captions in prompts, and employing soft labels for student training. These choices together yield the highest robustness and accuracy in TCKR.

\end{document}